\documentclass[10pt,twocolumn,letterpaper]{article}

\usepackage[pagenumbers]{cvpr} 

\usepackage[T1]{fontenc} 

\usepackage{cuted}    
\usepackage{currfile} 
\usepackage{caption}  
\renewcommand{\topfraction}{0.90}
\renewcommand{\dbltopfraction}{0.92}
\renewcommand{\textfraction}{0.06}
\renewcommand{\floatpagefraction}{0.88}
\renewcommand{\dblfloatpagefraction}{0.88}
\makeatletter
\renewcommand{\paragraph}{\@startsection{paragraph}{4}{\z@}{.29em}{-0.55em}{\normalfont\normalsize\bfseries}}
\g@addto@macro\normalsize{%
  \setlength{\abovedisplayskip}{4.5pt plus 1.5pt minus 1.5pt}%
  \setlength{\belowdisplayskip}{4.5pt plus 1.5pt minus 1.5pt}%
  \setlength{\abovedisplayshortskip}{2.5pt plus 1pt minus 1pt}%
  \setlength{\belowdisplayshortskip}{2.5pt plus 1pt minus 1pt}%
}
\makeatother

\usepackage{microtype}
\usepackage{booktabs}
\usepackage{colortbl}
\usepackage{multirow}
\usepackage{makecell}
\usepackage{tabularx}
\newcolumntype{Y}{>{\centering\arraybackslash}X}
\usepackage{bm}
\usepackage{amsthm}
\usepackage{pifont}
\usepackage{xspace}
\usepackage[most]{tcolorbox}
\usepackage{tikz}
\usetikzlibrary{arrows.meta,positioning,calc,fit,backgrounds,shapes.misc}
\usepackage{algorithm}
\usepackage{algpseudocode}

\newcommand{\method}{HinT\xspace}
\newcommand{\cmark}{\ding{51}}

\newcommand{\vel}{\bm{v}}
\newcommand{\rt}{\tilde{r}}
\newcommand{\habit}[1]{{\color{orange!80!black}#1}}
\newcommand{\fixw}[1]{{\color{green!45!black}\textbf{#1}}}

\definecolor{oursbg}{HTML}{F0EBF7}
\definecolor{basebg}{HTML}{F2F3F7}
\definecolor{sdgray}{HTML}{747184}
\newcommand{\best}[1]{\textbf{#1}}

\newcommand{\sd}[1]{{\scriptsize\textcolor{sdgray}{\,$\pm$\,#1}}}
\newcommand{\sdz}[1]{\phantom{\sd{#1}}}
\newcommand{\bestpm}[2]{\textbf{#1}{\scriptsize\textcolor{sdgray}{\,$\pm$\,#2}}}

\usepackage{environ}
\newsavebox{\fittabBox}
\newlength{\fittabWa}\newlength{\fittabWb}\newlength{\fittabTa}\newlength{\fittabTb}\newlength{\fittabT}
\newcounter{fittabIter}
\newif\iffittabGo
\newcommand{\fittabMeasure}[1]{\sbox{\fittabBox}{\begin{tabular}{#1}\BODY\end{tabular}}}
\NewEnviron{fittab}[2]{%
  \setlength{\fittabTa}{0pt}\setlength{\tabcolsep}{\fittabTa}\fittabMeasure{#2}\setlength{\fittabWa}{\wd\fittabBox}%
  \setlength{\fittabTb}{1pt}\setlength{\tabcolsep}{\fittabTb}\fittabMeasure{#2}\setlength{\fittabWb}{\wd\fittabBox}%
  \setcounter{fittabIter}{0}\fittabGotrue
  \loop
    \ifdim\fittabWb=\fittabWa \fittabGofalse \fi
    \ifdim\dimexpr#1-\fittabWb\relax<0.02pt \ifdim\dimexpr\fittabWb-#1\relax<0.02pt \fittabGofalse \fi\fi
    \ifnum\value{fittabIter}>9 \fittabGofalse \fi
  \iffittabGo
    \stepcounter{fittabIter}%
    \setlength{\fittabT}{\dimexpr\fittabTb+(\dimexpr#1\relax-\fittabWb)*\number\dimexpr\fittabTb-\fittabTa\relax/\number\dimexpr\fittabWb-\fittabWa\relax\relax}%
    \ifdim\fittabT<0pt \setlength{\fittabT}{0pt}\fi
    \setlength{\fittabTa}{\fittabTb}\setlength{\fittabWa}{\fittabWb}%
    \setlength{\fittabTb}{\fittabT}\setlength{\tabcolsep}{\fittabTb}\fittabMeasure{#2}\setlength{\fittabWb}{\wd\fittabBox}%
  \repeat
  \setlength{\tabcolsep}{\fittabTb}%
  \typeout{fittab: target \the\dimexpr#1\relax\space width \the\fittabWb\space tabcolsep \the\tabcolsep\space iters \arabic{fittabIter}}%
  \begin{tabular}{#2}\BODY\end{tabular}%
}
\newcolumntype{F}[1]{>{\columncolor{white}[0pt][\tabcolsep]}#1}
\newcolumntype{E}[1]{>{\columncolor{white}[\tabcolsep][0pt]}#1}

\AtEndPreamble{%
  \crefname{appendix}{Appendix}{Appendices}%
  \Crefname{appendix}{Appendix}{Appendices}%
  \crefname{algorithm}{Alg.}{Algs.}%
  \Crefname{algorithm}{Algorithm}{Algorithms}%
}

\definecolor{cvprblue}{rgb}{0.21,0.49,0.74}
\usepackage[pagebackref,breaklinks,colorlinks,allcolors=cvprblue]{hyperref}

\def\confName{CVPR}
\def\confYear{2027}

\title{Distill the Correction, Not the Rewrite:\\Hindsight Twins for Self-Improving Text-to-Image Models}

\newcommand\blfootnote[1]{%
  \begingroup
  \renewcommand\thefootnote{}\footnote{#1}%
  \addtocounter{footnote}{-1}%
  \endgroup
}
\author{
Jiaze Li$^{1}$\quad
Jingyang Chen$^{2}$\quad
Wenhao Zhao$^{3}$\quad
Haoran Xu$^{4}$\\
Zhenlong Yuan$^{1}$\quad
Dianmo Sheng$^{1}$\quad
Yuanxing Zhang$^{1,*}$\\[3pt]
{\normalsize $^{1}$KlingAI Research\quad
$^{2}$Fudan University\quad
$^{3}$South China Normal University\quad
$^{4}$Zhejiang University}\vspace{-6pt}
}

\begin{document}
\maketitle
\begin{strip}
\centering
\captionsetup{hypcap=false}
\includegraphics[width=0.825\linewidth,trim={1bp 21bp 24bp 8bp},clip]{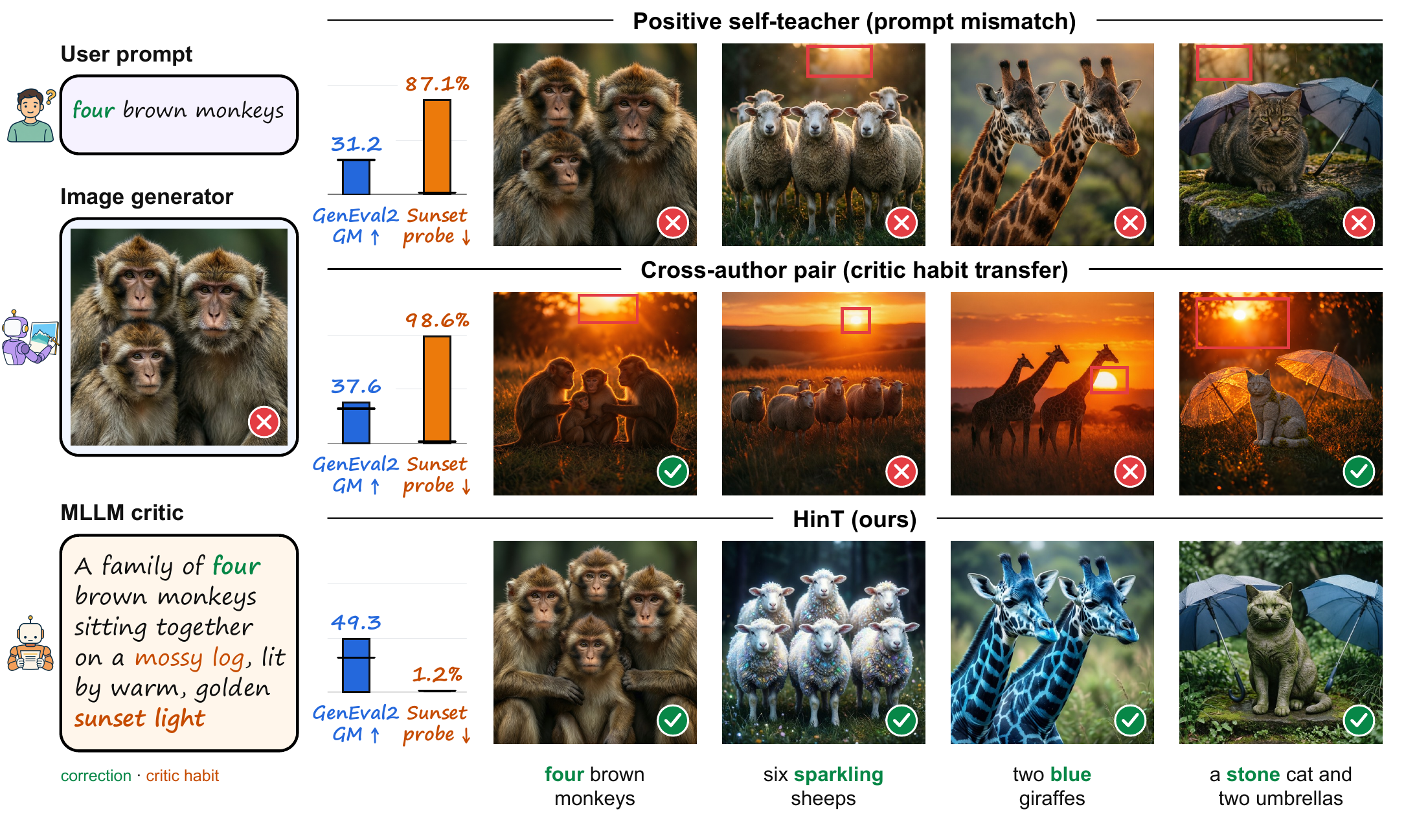}\par
\captionof{figure}{\textbf{Distill the correction, not the rewrite.} \textbf{Left:} Given an image depicting three monkeys in response to the prompt ``four brown monkeys,'' the MLLM critic corrects the count in its revision (\fixw{green}) but rewrites the scene with unrequested details, such as a mossy log and sunset lighting (\habit{orange}). \textbf{Right:} Qwen-Image-2512 after four rounds of self-distillation guided by this sunset-favoring critic. Both the positive self-teacher and the cross-author pair absorb the critic's sunset bias (red boxes) or fail to satisfy the prompt requirements, whereas HinT isolates the intended correction while keeping sunset prevalence close to that of the base model. Bars report GenEval2 Soft-TIFA geometric mean (GM~$\uparrow$) and sunset-probe rate ($\downarrow$) from Appendix~\cref{tab:habit}, with black ticks marking the base model.}
\label{fig:teaser}
\end{strip}

\begin{abstract}\blfootnote{$^*$Corresponding author.}
Text-to-image models can learn from their own mistakes by distilling a self-teacher conditioned on a vision-language critic's revised prompt. However, such self-teachers barely improve alignment, and both they and their hindsight-contrast variants absorb the critic's unintended additions to the scene, such as lighting or scenery. We attribute this to two factors: (1) the prompt already specifies the desired fix, so a useful revision must go beyond the failed span, often rewriting the entire scene and adding details that overshadow the fix; (2) although the critic localizes the error to a single span, self-teachers condition on the rewrite, entangling the intended fix with unintended additions. Motivated by these insights, we propose Hindsight Twins (HinT), which fills the critic's revision template once with the intended fix and once with the observed error and distills only their velocity difference, anchored at the user's prompt. Because text encoders spread the edit of a single slot to all other tokens, HinT further aligns the twins' token states outside the slot. Across Qwen-Image-2512, SD3.5-M, and FLUX.2-klein, HinT keeps the rate of critic-induced bias at the base-model level while matching or exceeding the alignment gains of prior self-distillation, and it keeps improving over twelve rounds of self-training where contrastive baselines decay.
\end{abstract}

\section{Introduction}
\label{sec:intro}


While a vision-language model (VLM) can readily diagnose generation errors such as missing objects or wrong counts~\cite{han2026unicorn,jin2025srum,hao2026dreaming}, turning its critique into weight updates remains difficult. Applying reinforcement learning (RL)~\cite{black2024ddpo,liu2025flowgrpo} collapses each critique into a scalar reward, discarding which requirement failed and how to fix it. Group-based online RL~\cite{liu2025flowgrpo,zheng2025diffusionnft} also needs multiple rollouts per prompt for advantage estimation, failing under a single-rollout budget in \cref{sec:exp}. To preserve verbal feedback in one rollout~\cite{agarwal2024gkd,li2026video}, on-policy self-distillation (OPSD)~\cite{zhao2026opsd,hubotter2026sdpo} supervises failed student samples with a teacher conditioned on the critic's revised prompt~\cite{vapnik2009lupi,snell2022context,lin2026peopsd,wu2026unievo,wang2026dopcd}: the \emph{positive self-teacher} distills this revision directly, whereas the \emph{cross-author pair} contrasts it with a hindsight-edited user prompt to push away from the error in the failed image.


However, both designs assume that a revision modifies only the diagnosed error. In practice, asked for a complete, more explicit prompt~\cite{wu2026unievo,lin2026peopsd,wang2026dopcd}, the critic rewrites the whole scene in its own style. When ``four brown monkeys'' yields only three in \cref{fig:teaser}, for instance, it restores the count (``four'') within an unrequested scene: ``A family of four brown monkeys sitting together on a mossy log, lit by warm, golden sunset light.'' In language modeling, contrastive self-distillation cancels such style drift by placing positive and negative solutions in a shared template~\cite{pan2026rlcsd,baumann2026repulsive}, whereas the cross-author pair contrasts this free-form description against a minimal user-prompt edit (``three brown monkeys''), leaving the rewrite intact. As \cref{subsec:fingerprints} shows, the positive self-teacher barely improves alignment, and both baselines absorb what the critic adds, from unrequested scenery to sunset lighting in over 82\% of test images under a sunset-favoring critic. Yet replacing only the failed span in the user's prompt is no option either: since the prompt already states the desired fix, it stays unchanged and yields no training signal. Thus, we attribute this dilemma to two factors:

\begin{tcolorbox}[enhanced, colback=gray!4, colframe=gray!45, boxrule=0.4pt, arc=1.5pt, left=1.2mm, right=1.2mm,
  top=0.8mm, bottom=0.8mm, before skip=4pt, after skip=4pt]
\begin{itemize}[leftmargin=1.0em, itemsep=2pt, topsep=0pt]
  \item \textit{Because replacing only the failed span leaves the original prompt unchanged, a meaningful revision must expand the prompt; in doing so, the critic rewrites the entire scene, introducing unrequested setting, lighting, and phrasing that far outweigh the fix.}
  \item \textit{Although the critic can localize the error to a specific span, existing self-teachers condition on the entire rewritten prompt without controlling for these additional changes, so their supervision entangles the intended fix with the critic's unrequested additions.}
\end{itemize}
\end{tcolorbox}

Motivated by these insights, we propose Hindsight Twins (HinT), which restores span-level localization inside the critic's own wording. Alongside the observed error and the intended fix, the critic returns its revision as a slotted template; filling the slot with the fix yields the revised prompt, whereas filling it with the observed error yields its \emph{hindsight twin}, a critic-worded description that the failed sample already satisfies~\cite{andrychowicz2017her,zhang2023hir}. By comparison, contrasting the user's prompt against its hindsight edit is also span-local, but it lacks scene context and trains unstably in \cref{sec:exp}. Because text encoders propagate a single-slot edit to surrounding tokens, HinT further aligns the twin's token states outside the slot to those of the revision in \cref{subsec:align}. With the two prompts matched outside the slot, adding the teacher velocity difference between the two prompts to the original-prompt anchor cancels the shared rewrite and keeps only the fix in \cref{fig:method}, so refreshing the teacher from the student each round compounds the correction rather than the critic's habits.\looseness=-1

Across Qwen-Image-2512, SD3.5-M, and FLUX.2-klein, HinT consistently improves compositional alignment without inheriting the critic's visual habits, and sustains its gains over multi-round training. On held-out GenEval2 prompts, it raises EMA Soft-TIFA geometric mean (GM) over the base model by 11.4, 11.1, and 1.9 points (2.8 current; 7.9 EMA at round twelve) on the three backbones, respectively; compared with the cross-author pair, it leads by 4.4 and 1.6 points on Qwen-Image-2512 and SD3.5-M, and overtakes it on FLUX.2-klein as the pair decays over rounds, while keeping habit rates at the base-model level under sunset, cold-light, and vintage critics. Extended analysis in \cref{sec:analysis} further validates our diagnosis: adding the rewrite back into HinT's target reintroduces the critic's habit without improving alignment, forbidding the critic from expanding the prompt hurts contrastive gains, and aligning token states keeps critic-added scenery out of the contrast.

In summary, our contributions are threefold:
\begin{itemize}[leftmargin=1.2em, itemsep=0.5pt, topsep=1pt, parsep=0pt]
    \item We diagnose why text-to-image self-distillation inherits the critic's style: through a velocity-space decomposition and controlled habit probes, we show that the critic's scene rewrite dominates the teacher signal and compounds into a student habit that cross-author contrasts fail to cancel.
    \item We propose Hindsight Twins, which fills a slotted critic revision with the intended fix and the observed error, distilling their velocity difference under an evolving teacher.\looseness=-1
    \item We introduce token-state alignment, which keeps the twins identical outside the slot in feature space, preventing critic-added scenery from leaking into the contrast.
\end{itemize}

\section{Motivation}
\label{sec:motivation}

\begin{figure*}[t]
  \centering
  \includegraphics[width=\linewidth,trim={0 3bp 0 6bp},clip]{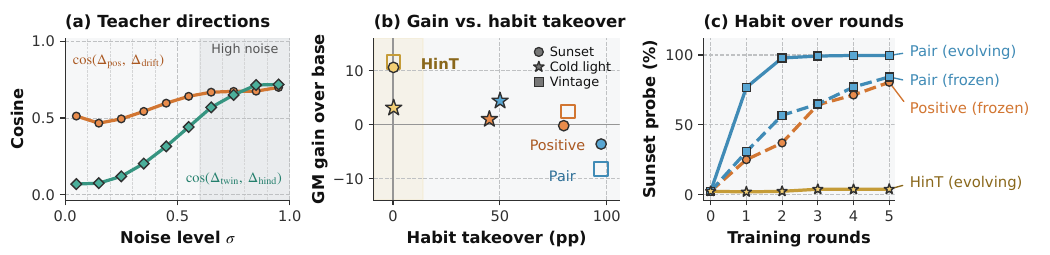}
  \caption{\textbf{Rewrite drift dominates the teacher signal and becomes a student habit on SD3.5-M.} (a) Across 1,146 re-noised failures from \cref{eq:decomp}, the full-revision update aligns with rewrite drift ($\cos(\Delta_{\rm pos},\Delta_{\rm drift})$), whereas the twin and user-prompt local fixes align at high noise ($\cos(\Delta_{\rm twin},\Delta_{\rm hind})$, shaded). (b) Under sunset (circles), cold-light (stars), and vintage-film (squares) critics, both the positive self-teacher (\emph{Positive}) and cross-author pair (\emph{Pair}) suffer large habit takeover with minimal or negative GM gain, whereas \method (ours) raises GM near zero takeover. (c) Sunset rates on student rollouts exceed 80\% by round five under frozen teachers (\emph{frozen}) and reach 98\% within two rounds when the \emph{Pair} teacher is updated from the student each round (\emph{evolving}), whereas \method stays below 4\%.}
  \label{fig:motivation}
\end{figure*}

\subsection{Self-Teachers Inherit the Critic's Rewrite}
\label{subsec:fingerprints}

On-policy self-distillation uses a critic's revised prompt to guide the student whenever a generated image misses a requirement. Because the user's prompt already states that requirement, replacing only the failed span with the intended fix leaves the prompt unchanged and provides no training signal. Prior self-teachers~\cite{wu2026unievo,lin2026peopsd,wang2026dopcd} therefore ask the critic to expand the prompt into a complete, more explicit scene description. However, expanding the prompt causes the critic to rewrite the entire scene in its own style. As detailed in \cref{sec:supp_critic}, this behavior is pervasive across our experiments, where the default Qwen3-VL-32B critic expands 10-word median prompts into 22 to 25-word descriptions and adds unrequested setting or lighting in 59\% to 74\% of revisions.

Our distillation experiments show that training on these expanded revisions causes the student to internalize the critic's scene preferences as visual habits. Under the default critic, both the positive self-teacher and the cross-author pair replace plain backgrounds with the critic's added scenery in Appendix~\cref{tab:fingerprint}, while the positive self-teacher barely improves alignment. We further test this habit transfer in a controlled setting by instructing the critic to include warm sunset light, cold light, or vintage film in every revised scene while keeping its error diagnosis unchanged. Both baselines readily absorb each injected habit across backbones: on SD3.5-M in \cref{fig:motivation}b, they incur 45 to 97 percentage points of habit takeover while yielding minimal or negative GM gains. Over training rounds in \cref{fig:motivation}c, sunset prevalence under both baselines jumps from 2.2\% to over 80\% with a frozen teacher, and reaches nearly 100\% within two rounds for the cross-author pair when its teacher is updated from the student each round. Prompting the critic to restrain its rewrite~\cite{wu2026unievo} fails to block injected habits, while forcing minimal edits sacrifices alignment gains, as shown in Appendix~\cref{tab:critic_style}.


From the above observations, we summarize two factors behind this phenomenon: (1) \textit{the critic's unrequested scene rewrite outweighs the single-span fix}. (2) \textit{conditioning on the full revision without a matched template lets the surrounding rewrite enter the teacher's signal alongside the fix.}\looseness=-1

\begin{figure*}[t]
  \centering
  \includegraphics[width=\linewidth,trim={13bp 3bp 8bp 12bp},clip]{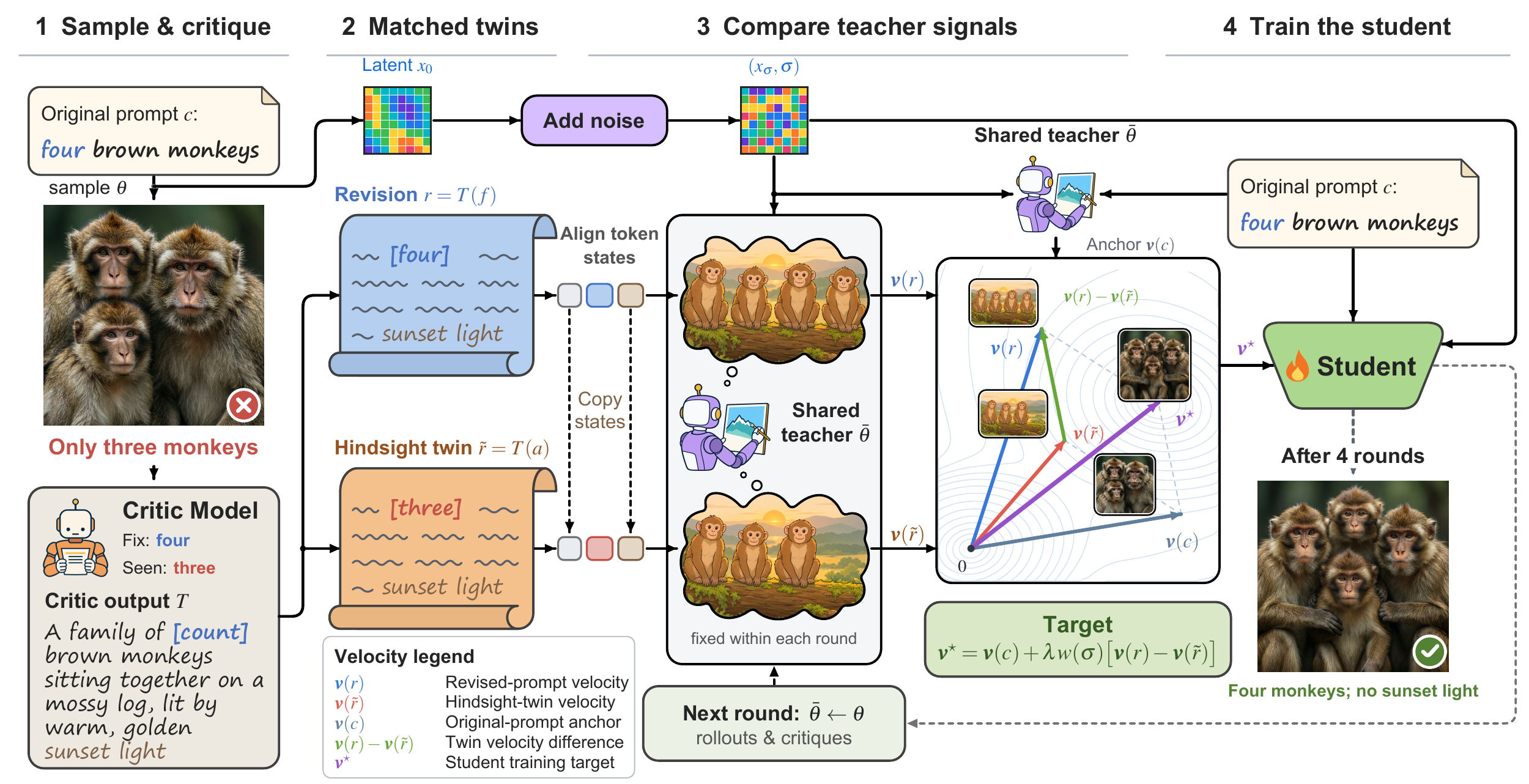}
  \caption{\textbf{Overview of Hindsight Twins (\method).}
  \textbf{(1)} On a failed student rollout, the critic reports the intended fix $f$ (\emph{Fix}), the observed error $a$ (\emph{Seen}), and a slotted revision template $T$. \textbf{(2)} Filling $T$ gives $r=T(f)$ and its twin $\rt=T(a)$, which differ only in the slot; their token states are aligned outside it as described in \cref{subsec:align}. \textbf{(3)} At a shared re-noised latent $(x_\sigma,\sigma)$, the round-start teacher $\bar\theta$ adds the twin velocity difference, scaled by $\lambda w(\sigma)$, to its original-prompt anchor (the thought bubbles illustrate the two conditions, not sampled images). \textbf{(4)} Only the student's LoRA weights are trained toward this detached target under $c$ via \cref{eq:target}, and the teacher is refreshed from the student every round. Photographs: Qwen-Image-2512 before and after four \method rounds under the sunset-favoring critic (same prompt and seed).\looseness=-1}
  \label{fig:method}
\end{figure*}

\subsection{Localizing the Correction Is Necessary}
\label{subsec:drift}
\label{subsec:decomp}

To verify both factors in \cref{subsec:fingerprints}, we separate the local fix from the scene rewrite in the teacher's velocity field. Given a failed image for user prompt $c$ and the critic's expanded revision $r$, replacing the failed span with what the image actually shows yields two descriptions of the flawed image: $h$ in the user's words, and a matched counterpart $\rt$ inside the critic's scene description. Let $\vel(p)$ denote the teacher's predicted velocity under prompt $p$ on the same re-noised image at noise level $\sigma$. Supervising a student conditioned on $c$ toward $\vel(r)$ applies the full-prompt update $\Delta_{\rm pos}=\vel(r)-\vel(c)$, which decomposes into three terms:
\begin{equation}
  \Delta_{\rm pos}=\underbrace{\vel(r)\!-\!\vel(\rt)}_{\Delta_{\rm twin}}
  -\underbrace{\big(\vel(c)\!-\!\vel(h)\big)}_{\Delta_{\rm hind}}+\underbrace{\vel(\rt)\!-\!\vel(h)}_{\Delta_{\rm drift}}.
  \label{eq:decomp}
\end{equation}
Here, $\Delta_{\rm twin}$ and $\Delta_{\rm hind}$ change only the failed span while keeping the critic's or the user's surrounding text fixed, whereas $\Delta_{\rm drift}$ compares two descriptions of the same flawed image and captures the critic's scene rewrite alone. Because $c$ and $r$ both state the intended fix, $\Delta_{\rm twin}$ and $\Delta_{\rm hind}$ enter \cref{eq:decomp} with opposite signs and largely cancel out, leaving $\Delta_{\rm pos}$ dominated by $\Delta_{\rm drift}$. Meanwhile, the cross-author pair uses $\vel(r)-\vel(h)=\Delta_{\rm twin}+\Delta_{\rm drift}$: subtracting $h$ recovers the local fix $\Delta_{\rm twin}$, but still leaves the entire scene rewrite $\Delta_{\rm drift}$ in the teacher signal.

To test this decomposition in practice, we probe the SD3.5-M teacher across 1,146 re-noised first-round failures from $\sigma=0.05$ to $0.95$ in \cref{fig:motivation}a, matching results on Qwen-Image-2512 and FLUX.2-klein in Appendix~\cref{tab:probe_qwen,tab:probe_flux}. Across the entire noise schedule, the full-prompt update $\Delta_{\rm pos}$ closely tracks the rewrite drift $\Delta_{\rm drift}$ with a median cosine of 0.47 to 0.70; at high noise ($\sigma\ge0.65$), where global layout forms~\cite{jeong2025diffusionclassifiers}, only 5.7\% of the energy of $\Delta_{\rm pos}$ projects onto the local fix $\Delta_{\rm twin}$. In contrast, when the difference is restricted to the failed span, $\Delta_{\rm twin}$ and $\Delta_{\rm hind}$ align strongly at high noise despite different surrounding wording, with their median cosine rising from at most 0.20 for $\sigma\le0.35$ to 0.65-0.72 for $\sigma\ge0.75$. These results substantiate our attribution in \cref{subsec:fingerprints} and establish two key insights: (1) \textit{whole-prompt self-distillation is overwhelmingly driven by the critic's unrequested scene rewrite rather than the fix}; and (2) \textit{localizing the teacher contrast to the failed span within a shared scene description isolates the intended correction at high noise while canceling rewrite drift.}

\section{Method}
\label{sec:method}

We first review rectified-flow self-distillation in \cref{subsec:prelim}. We then present Hindsight Twins (\method) in \cref{fig:method}, which separates the diagnosed fix from the critic's rewrite in three steps: constructing matched twin prompts from a slotted critique in \cref{subsec:twins}, aligning their token states outside the edited slot in \cref{subsec:align}, and compounding the resulting update across rounds with an evolving teacher in \cref{subsec:compound}.

\subsection{Preliminaries}
\label{subsec:prelim}

Given a clean latent $x_0$, Gaussian noise $\epsilon\sim\mathcal N(0,I)$, and a noise level $\sigma\in[0,1]$, rectified-flow transformers~\cite{liu2023rectified,lipman2023flow,esser2024sd3} form $x_\sigma=(1-\sigma)x_0+\sigma\epsilon$ and train a velocity network $\vel_\theta(x_\sigma,\sigma,p)$ conditioned on prompt $p$ to predict $\epsilon-x_0$. In on-policy self-distillation, a critic evaluates each student rollout $x_0\sim p_\theta(\cdot\mid c)$; when a generation fails, the critic's revised prompt $r$ provides privileged context~\cite{lin2026peopsd,wu2026unievo,wang2026dopcd}, and a teacher $\bar\theta$ conditioned on $r$ supervises the student on its own re-noised latent:
\begin{equation}
  \mathcal{L}_{\rm pos}(\theta)=\mathbb{E}_{\sigma,\epsilon}\,\big\|\vel_\theta(x_\sigma,\sigma,c)-
  \mathrm{sg}\big[\vel_{\bar\theta}(x_\sigma,\sigma,r)\big]\big\|^2,
  \label{eq:pos}
\end{equation}
where $\mathrm{sg}[\cdot]$ stops gradients and $\bar\theta$ is a frozen or slowly updated copy of the student. Because $\vel_{\bar\theta}(x_\sigma,\sigma,r)=\vel_{\bar\theta}(x_\sigma,\sigma,c)+\Delta_{\rm pos}$, minimizing \cref{eq:pos} drives the student's prediction along $\Delta_{\rm pos}$, which is dominated by the scene rewrite $\Delta_{\rm drift}$ in \cref{eq:decomp}.

\subsection{Hindsight Twins}
\label{subsec:twins}

\paragraph{Slotted critique.} To pair the revised prompt with a negative counterpart that shares the critic's exact scene description, we ask the critic for a slotted revision template $T$ alongside its error diagnosis, as detailed in \cref{sec:supp_critic}. For a failed rollout in \cref{fig:method}, the critic identifies the intended fix $f$, the observed error $a$ in place of the failed span in $c$, and a slotted revision template $T$ at the diagnosed span. Filling the slot with $f$ and $a$ yields the revised prompt $r=T(f)$ and its hindsight twin $\rt=T(a)$. Unlike the user-worded hindsight edit $h$, $r$ and $\rt$ share every word outside the slot, so their contrast cancels the surrounding rewrite even when the critic heavily expands the prompt, as verified in Appendix~\cref{tab:rewrite_buckets}.

\paragraph{Correction-only target.} To supervise the student under $c$ without pulling it toward the critic's rewrite, \method replaces $\vel_{\bar\theta}(x_\sigma,\sigma,r)$ in \cref{eq:pos} by adding the twin velocity difference to the original-prompt anchor, see \cref{fig:method}. Writing $\vel_{\bar\theta}(p)$ for $\vel_{\bar\theta}(x_\sigma,\sigma,p)$ at a shared $(x_\sigma,\sigma)$ with $\sigma\sim\mathcal U(0.05,0.95)$, the target velocity on a failed rollout is
\begin{equation}
  \vel^\star=\vel_{\bar\theta}(c)+\lambda\,w(\sigma)\,\big[\vel_{\bar\theta}(r)-\vel_{\bar\theta}(\rt)\big],
  \label{eq:target}
\end{equation}
where $\lambda>0$ scales the contrast and $w(\sigma)=2\sigma$. As shown in \cref{subsec:drift}, $\Delta_{\rm twin}$ reliably isolates the intended fix at high noise, where global layout forms, whereas its direction is unreliable at low noise once layout is already set. Setting $w(\sigma)=2\sigma$ keeps the average weight at $\mathbb{E}_\sigma[w(\sigma)]=1$ while attenuating the contrast for $\sigma<0.5$ and amplifying it for $\sigma>0.5$, concentrating corrective updates on high-noise states without changing the overall scale $\lambda$. On accepted rollouts, we omit the contrast term so that $\vel^\star=\vel_{\bar\theta}(c)$ anchors the student to the teacher, and train the student by substituting $\vel^\star$ for $\vel_{\bar\theta}(x_\sigma,\sigma,r)$ in \cref{eq:pos}. Because $r$ and $\rt$ differ only in the slot, canceling the shared rewrite leaves $\Delta_{\rm twin}$ with a much smaller norm than the full-prompt update $\Delta_{\rm pos}$ in Appendix~\cref{tab:probe_qwen,tab:probe_flux}. Eliminating $\Delta_{\rm drift}$ allows us to amplify this clean, span-local signal with $\lambda\ge1$~\cite{yang2026gopd,xu2026sg,lin2026dreopd} for effective correction without magnifying style drift.

\subsection{Token-State Alignment}
\label{subsec:align}

Sharing the template $T$ matches $r$ and $\rt$ in text space, yet encoding the two prompts separately still changes token representations outside the slot. Through bidirectional or causal attention, the slot edit propagates to surrounding tokens and lets critic-added scenery leak back into the velocity contrast in \cref{subsec:local}. To keep non-edited tokens identical in feature space, we copy $r$'s text-encoder output states onto $\rt$ at all positions outside the edited span in \cref{fig:method}, which replaces both prefix and suffix states under bidirectional encoders (\emph{span replacement}) and reduces to copying the shared suffix under causal encoders (\emph{suffix alignment}).

\subsection{Compounding the Correction}
\label{subsec:compound}

Keeping the teacher fixed at the base model $\theta_0$ caps how much the student can improve, since every update still anchors the student to the initial model. To accumulate gains across rounds, \method updates the teacher to the current student ($\bar\theta\gets\theta$) at the start of each round, samples a fresh set of rollouts for the critic to judge, and maintains an exponential moving average (EMA) of the student's LoRA weights~\cite{hu2022lora,tarvainen2017meanteacher}, as summarized in Appendix~\cref{alg:hint}. Refreshing the teacher, however, compounds whatever signal is present in the target. When the target retains the rewrite drift $\Delta_{\rm drift}$, each new teacher inherits the critic's style from the previous round and amplifies it further, as shown in \cref{fig:motivation}c. Because \method cancels the shared rewrite in $\Delta_{\rm twin}$ and anchors accepted rollouts at $\vel_{\bar\theta}(c)$, each teacher refresh carries forward only the learned corrections.

\section{Experiments}
\label{sec:exp}

We first describe our experimental setup in \cref{subsec:setup}. \cref{subsec:main} then compares \method with prior baselines across three backbones on held-out and disjoint-prompt evaluations, zero-shot transfer, and multi-round training. Next, \cref{subsec:habit} evaluates all methods under habit-carrying critics, and \cref{subsec:ablation} ablates the key components of \method. Qualitative comparisons are provided in Appendix~\cref{fig:habitgrid_qwen,fig:qualitative,fig:vintage}.

\subsection{Setup}
\label{subsec:setup}

\paragraph{Backbones and protocols.} We train LoRA adapters~\cite{hu2022lora} on three rectified-flow backbones with distinct text encoders: Qwen-Image-2512~\cite{wu2025qwenimage} (Qwen2.5-VL), SD3.5-M~\cite{esser2024sd3} (CLIP+T5), and FLUX.2-klein-base-4B~\cite{bfl2025flux2} (Qwen3), with full settings in Appendix~\cref{tab:supp_hparams}. Our primary protocol splits the 800 GenEval2 prompts~\cite{kamath2025geneval2} into 400 training and 400 held-out prompts (four images per prompt), training for four rounds of 60 updates on Qwen-Image-2512 and six rounds on SD3.5-M and FLUX.2-klein. We report EMA (decay 0.98) and final-round (Cur.) weights, and evaluate the checkpoints zero-shot on GenEval~\cite{ghosh2023geneval} and OneIG-Bench~\cite{chang2025oneig} in \cref{tab:ood}. Following~\cite{wu2026unievo}, we also evaluate Qwen-Image-2512 in \cref{tab:qwen} on 800 GenEval2 prompts after training on 800 disjoint grammar prompts detailed in \cref{sec:supp_impl}.\looseness=-1

\paragraph{Baselines.} Our baselines isolate the role of each component in \method.
The \emph{positive self-teacher}~\cite{lin2026peopsd,wu2026unievo,wang2026dopcd} supervises the student directly with $\vel_{\bar\theta}(r)$ via \cref{eq:pos}, distilling the critic's full revision without filtering.
To subtract what the failed image already shows, the \emph{cross-author pair} contrasts the full revision $r$ against the user-worded hindsight prompt $h$, adding $\vel_{\bar\theta}(r)-\vel_{\bar\theta}(h)$ to $\vel_{\bar\theta}(c)$.
The \emph{user-worded hindsight contrast} instead contrasts $c$ against $h$ directly in the user's words, isolating a local edit without the critic's scene context.
Rather than operating in velocity space, \emph{SFT on the revised prompt} applies $r$ at the sample level: it regenerates failed images from $r$, keeps those that pass the critic under $c$, and fine-tunes on them with flow matching.
Finally, \emph{DiffusionNFT}~\cite{zheng2025diffusionnft} replaces verbal critiques with a scalar reward (the fraction of satisfied questions per rollout) for on-policy RL.

\begin{table*}[t]
\centering
\footnotesize
\renewcommand{\arraystretch}{1.00}
\caption{\textbf{Comparison on held-out GenEval2 prompts.} We report Soft-TIFA GM under EMA and current (Cur.) weights alongside the independent InternVL3.5-38B judge score (J2); baseline implementation notes and seed standard deviations are in \cref{sec:supp_impl} and Appendix~\cref{tab:heldout_full}. \textbf{Bold} marks the best result per column.}
\label{tab:heldout}
\begin{fittab}{\textwidth}{@{}F{l} ccc ccc ccE{c}@{}}
\toprule
& \multicolumn{3}{c}{Qwen-Image-2512} & \multicolumn{3}{c}{SD3.5-M} & \multicolumn{3}{c}{FLUX.2-klein} \\
\cmidrule(lr){2-4}\cmidrule(lr){5-7}\cmidrule(l){8-10}
Method & EMA & Cur. & J2 & EMA & Cur. & J2 & EMA & Cur. & J2 \\
\midrule
\rowcolor{basebg}
Base model & \multicolumn{2}{c}{33.9} & 50.4 & \multicolumn{2}{c}{19.6} & 35.4 & \multicolumn{2}{c}{23.5} & 40.9 \\
Positive teacher & 33.5 & 34.8 & 51.5 & --- & 21.4 & 38.8 & --- & 20.0 & 36.5 \\
SFT on revised prompt & --- & 38.6 & 54.1 & --- & 21.8 & 39.2 & --- & 8.6 & 29.5 \\
DiffusionNFT & 33.3 & 33.8 & 50.4 & 19.8 & 20.0 & 36.0 & 23.0 & 22.5 & 40.5 \\
Cross-author pair & 40.9 & 41.9 & 56.5 & 29.1 & \best{28.5} & 45.8 & \best{27.3} & 22.4 & \best{45.8} \\
User hindsight & 43.2 & 42.0 & 57.2 & 24.8 & 24.7 & 41.1 & 10.4 & 5.4 & 25.8 \\
\rowcolor{oursbg}
\textbf{\method (ours)} & \best{45.3} & \best{44.7} & \best{59.2} & \best{30.7} & 27.5 & \best{47.9} & 25.4 & \best{26.3} & 43.9 \\
\bottomrule
\end{fittab}
\end{table*}

\paragraph{Critic and judges.} All methods share a Qwen3-VL-32B critic~\cite{qwen2025qwen3vl}. Following the official GenEval2 protocol, we report Soft-TIFA geometric mean (GM) from its Qwen3-VL-8B judge as our primary metric, alongside a cross-family InternVL3.5-38B judge (J2, also used for habit probing)~\cite{wang2025internvl35} and detector-based GenEval to rule out same-family bias. Moreover, \cref{sec:supp_critic,sec:supp_eval} confirm that our findings hold with an InternVL3.5-38B critic, official benchmark scorers, and a second habit rater ($0.98$ correlation), while \cref{subsec:noise} tests robustness to simulated critic errors.\looseness=-1

\paragraph{Fairness and cost.} All methods share the same prompts, critic, rollout budget, LoRA rank, and update schedule, with each baseline tuned over its own teacher rule, EMA, and contrast scale in \cref{sec:supp_impl}. Main results average three to four seeds for \method and the cross-author pair and two to three for other baselines, reusing fixed settings without retuning on the 800-prompt Qwen-Image-2512 benchmark or zero-shot transfers. On failed rollouts, \method matches the cross-author pair at three teacher passes, adding under 4\% training time over the positive teacher and zero inference cost in Appendix~\cref{tab:supp_hparams}.

\subsection{Main Results}
\label{subsec:main}
\label{subsec:heldout}

\begin{figure*}[t]
  \centering
  \includegraphics[width=\linewidth,trim={0 3bp 0 2bp},clip]{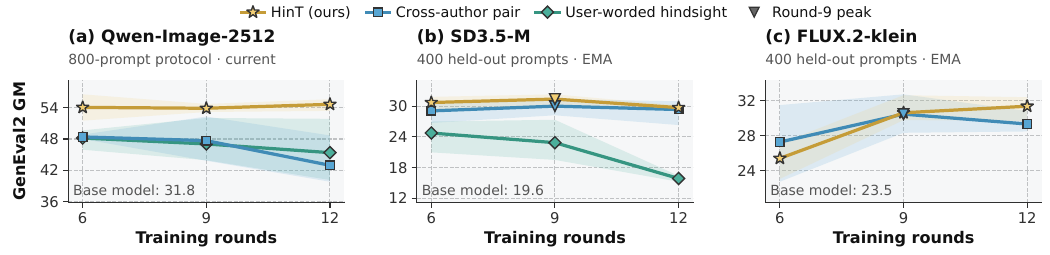}
  \caption{\textbf{Multi-round stability over twelve rounds.} Solid lines and shaded bands show seed means and ranges across rounds 6, 9, and 12, and triangles mark round-nine peaks, with full numbers in Appendix~\cref{tab:rounds}.}
  \label{fig:rounds}
\end{figure*}

\paragraph{GenEval2 alignment.}
As shown in \cref{tab:heldout}, \method reaches 45.3, 30.7, and 25.4 EMA Soft-TIFA GM (26.3 current on FLUX.2-klein) across Qwen-Image-2512, SD3.5-M, and FLUX.2-klein on held-out prompts, gaining 11.4, 11.1, and 1.9 points (2.8 current) over the base models while also improving ImageReward~\cite{xu2023imagereward} and HPSv2.1~\cite{wu2023hpsv2} in Appendix~\cref{tab:quality}.
On the 800-prompt Qwen-Image-2512 benchmark with disjoint training prompts in \cref{tab:qwen}, \method reaches 49.0 GM, surpassing the base model by 17.2 points.
Compared with full-revision baselines, \method beats the positive self-teacher by 5.4 to 11.8 GM on held-out prompts and 15.1 in \cref{tab:qwen}, and outperforms sample-level SFT by 6.7 to 16.8 GM, confirming that unfiltered revisions are dominated by rewrite drift $\Delta_{\rm drift}$.
Compared with the scalar-reward RL baseline DiffusionNFT, single-rollout \method leads by 21.6 points in \cref{tab:qwen} and by 11.0 over four-rollout DiffusionNFT in Appendix~\cref{tab:qwen_full}, showing that velocity distillation provides much denser supervision than a scalar reward.
Finally, \method outperforms the cross-author pair by 4.4 and 4.6 GM on Qwen-Image-2512 in \cref{tab:heldout,tab:qwen} and by 1.6 EMA GM (2.1 J2) on SD3.5-M; on FLUX.2-klein, where six-round EMA trails the higher-variance pair (25.4 vs.\ 27.3), \method achieves stronger current weights (26.3 vs.\ 22.4), overtakes it in later rounds in \cref{fig:rounds}c, and avoids the collapse of user hindsight (10.4 EMA GM), underscoring the value of canceling the critic's rewrite inside a shared scene template.

\begin{table}[t]
\centering
\footnotesize
\renewcommand{\arraystretch}{1.00}
\caption{\textbf{Evaluation on untouched benchmarks.} GE and OI denote GenEval overall score ($\times100$) and OneIG-Bench General-Object alignment, respectively, evaluated on the checkpoints from \cref{tab:heldout} (User hindsight is omitted on FLUX.2 where training collapsed). \textbf{Bold} indicates the best performance.}
\label{tab:ood}
\begin{fittab}{\linewidth}{@{}F{l} cc cc cE{c}@{}}
\toprule
& \multicolumn{2}{c}{Qwen-Image} & \multicolumn{2}{c}{SD3.5-M} & \multicolumn{2}{c}{FLUX.2} \\
\cmidrule(lr){2-3}\cmidrule(lr){4-5}\cmidrule(l){6-7}
Method & GE & OI & GE & OI & GE & OI \\
\midrule
\rowcolor{basebg}
Base model & 73.9 & 82.9 & 67.3 & 73.1 & 76.4 & 79.1 \\
Positive teacher & 74.3 & 82.9 & 68.9 & 74.6 & 72.7 & 77.5 \\
Cross-author pair & 75.2 & 83.4 & 73.5 & 75.6 & \best{79.6} & 79.9 \\
User hindsight & \best{77.9} & 83.0 & 72.1 & 72.6 & --- & --- \\
\rowcolor{oursbg}
\textbf{\method (ours)} & 77.3 & \best{84.1} & \best{74.1} & \best{76.1} & 77.5 & \best{80.0} \\
\bottomrule
\end{fittab}
\end{table}

\begin{table}[t]
\centering
\footnotesize
\renewcommand{\arraystretch}{1.00}
\caption{\textbf{GenEval2 results on Qwen-Image-2512 with disjoint training prompts.} Models train for four rounds on 800 disjoint grammar prompts and are evaluated on all 800 GenEval2 prompts under current weights (EMA reports GM under EMA weights), with full-protocol, multi-rollout, and test-time revision comparisons in Appendix~\cref{tab:qwen_full}.}
\label{tab:qwen}
\begin{fittab}{\linewidth}{@{}F{l} c c c c E{c}@{}}
\toprule
Method & GM $\uparrow$ & AM $\uparrow$ & J2 $\uparrow$ & Pick $\uparrow$ & EMA $\uparrow$ \\
\midrule
\rowcolor{basebg}
Base model & 31.8 & 78.3 & 48.8 & 23.86 & --- \\
DiffusionNFT & 27.4 & 76.4 & 44.6 & 23.78 & 30.4 \\
Positive teacher & 33.9 & 78.6 & 51.3 & 23.83 & 32.6 \\
Cross-author pair & 44.4 & 82.2 & 58.7 & 23.79 & 42.5 \\
User hindsight & 43.5 & 82.8 & 57.6 & 23.92 & 43.0 \\
\rowcolor{oursbg}
\textbf{\method (ours)} & \best{49.0} & \best{84.2} & \best{62.8} & \best{23.99} & \best{47.2} \\
\bottomrule
\end{fittab}
\end{table}

\paragraph{Zero-shot transfer.}
As shown in \cref{tab:ood}, \method improves over every base model on both untouched benchmarks, reaching 77.3, 74.1, and 77.5 on GenEval and achieving top OneIG-Bench scores of 84.1, 76.1, and 80.0 across Qwen-Image-2512, SD3.5-M, and FLUX.2-klein, while leading on all three backbones under the official OneIG scorer and improving T2I-CompBench++~\cite{huang2025t2icompbenchpp} in Appendix~\cref{tab:oneig_official,tab:t2i}.
Because GenEval shares short templated prompts with GenEval2, user hindsight edges out \method on Qwen-Image-2512 (77.9 vs.\ 77.3), yet fails to transfer to scene-rich OneIG-Bench prompts (83.0 vs.\ 82.9 base, degrading SD3.5-M to 72.6); similarly, although the cross-author pair starts higher on six-round FLUX.2-klein GenEval, continued training degrades it below the 76.4 base to 72.6 by round twelve while \method rises to 78.8 in \cref{sec:supp_results}, showing that isolating $\Delta_{\rm twin}$ yields durable, transferable binding.

\paragraph{Multi-round stability.}
As shown in \cref{fig:rounds} and Appendix~\cref{tab:rounds}, \method sustains or expands its gains over twelve rounds while contrastive baselines degrade as retained drift compounds.
User hindsight drops steadily after round six on Qwen-Image-2512 (48.2 to 45.4 GM) and falls below the 19.6 base model to 15.9 on SD3.5-M.
Meanwhile, the cross-author pair decays by 5.5 points on Qwen-Image-2512 (48.4 to 42.9 vs.\ \method's 54.1 to 54.7), trails \method at the round-nine peak on SD3.5-M (30.0 vs.\ 31.4 EMA) before both ease slightly at round twelve (29.3 vs.\ 29.7), and on FLUX.2-klein peaks at round nine (30.5 vs.\ \method's 30.6 EMA) before slipping to 29.3 while \method climbs to 31.4 EMA (31.3 vs.\ 27.2 current).

\begin{figure}[t]
  \centering
  \includegraphics[width=\linewidth,trim={0 0 0 4bp},clip]{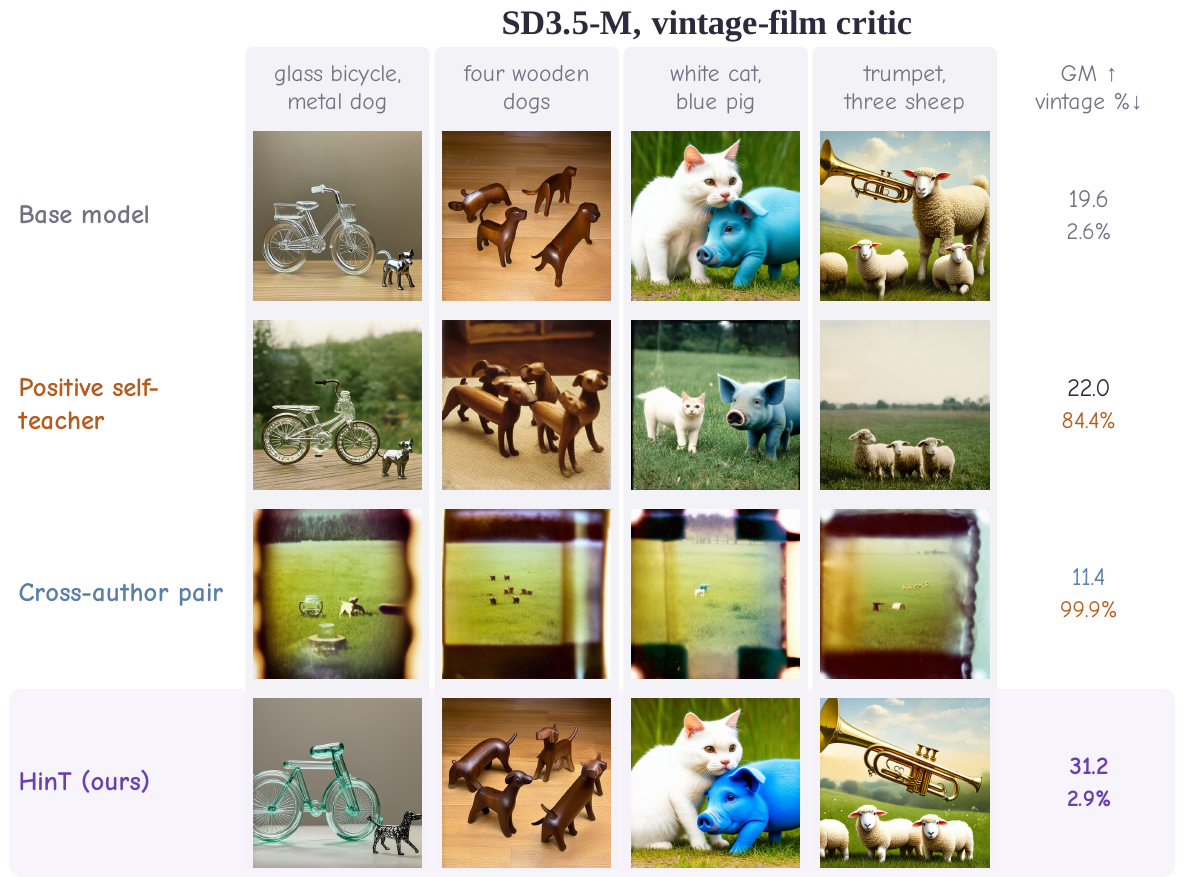}
  \caption{\textbf{Habit transfer on SD3.5-M under the vintage-film critic.} Columns share a test prompt and seed after six training rounds; the rightmost column reports three-seed mean GM and vintage prevalence from Appendix~\cref{tab:habit}. Unlike the baselines, \method fixes the prompt errors without absorbing the vintage style.}
  \label{fig:habitgrid}
\end{figure}

\subsection{Robustness to the Critic's Habits}
\label{subsec:habit}

\cref{sec:motivation} showed that both full-revision distillation and cross-author pairs retain the rewrite drift $\Delta_{\rm drift}$ in \cref{eq:decomp}, causing the student to inherit the critic's habits.
Beyond the default critic's natural scene additions analyzed in \cref{subsec:learns}, we run controlled stress tests across eight backbone--critic settings favoring warm sunset light, cold overcast light, or vintage film framing under current weights in \cref{fig:habitgrid} and Appendix~\cref{fig:habitgrid_qwen,tab:habit,tab:fingerprint}.
Under the sunset critic, \method keeps sunset prevalence at 1.2\% to 2.4\% across backbones (matching the 1.4\% to 2.7\% base rates) while raising GM from 31.8, 19.6, and 23.5 to 49.3, 30.2, and 25.4 on Qwen-Image-2512 (800-prompt protocol), SD3.5-M, and FLUX.2-klein.
In contrast, the positive self-teacher and cross-author pair drive sunset prevalence to 82.1\% to 99.6\%; even though Soft-TIFA does not penalize unrequested lighting, this style takeover costs the pair over half its alignment gain on Qwen-Image-2512 (37.6 vs.\ 44.5 GM under the default critic) and drops SD3.5-M below the base model (16.0 vs.\ 19.6 GM).
Across the cold-light and vintage-film critics, both baselines suffer 52.3\% to 99.9\% style takeover, corrupting nearly every image even where FLUX.2-klein GM remains higher, and dragging the SD3.5-M pair down to 11.4 GM at 99.9\% vintage prevalence in \cref{fig:habitgrid} and Appendix~\cref{fig:vintage}, while \method stays within 1.5 points of every base habit rate (2.9\% vintage and 31.2 GM on SD3.5-M) while improving GM in all eight settings.

\begin{table}[t]
\centering
\footnotesize
\renewcommand{\arraystretch}{1.00}
\caption{\textbf{Ablation study of \method's components.} Upper block: SD3.5-M on held-out prompts (base: 19.6 GM, 2.2\% Sun.), ablating target choice (Rows 1 to 3, with sunset rates Sun.\ in \%) and training schedule (Rows 4 to 8): token-state alignment (Align.), evolving teacher (Evol.), high-noise weighting $w(\sigma)=2\sigma$, and LoRA EMA. Lower block: suffix alignment on FLUX.2-klein under the sunset critic (base: 23.5 GM, 2.7\% Sun.), detailed in Appendix~\cref{tab:embedding}. \textbf{Bold} marks the best result per block.}
\label{tab:ablation}
\begin{fittab}{\linewidth}{@{}F{r} l cccc c E{c}@{}}
\toprule
\# & Target & Align. & Evol. & $w(\sigma)$ & EMA & GM $\uparrow$ & Sun. $\downarrow$ \\
\midrule
\multicolumn{8}{@{}l}{\textit{SD3.5-M (default critic for GM; sunset critic for Sun.)}} \\
1 & Positive self-teacher & --- & --- & --- & --- & 21.4 & 82.1 \\
2 & Cross-author pair & --- & --- & --- & --- & 25.4 & 81.9 \\
3 & Hindsight twins & \cmark & --- & --- & --- & 25.4 & \best{2.2} \\
\addlinespace[1.5pt]
4 & Hindsight twins & \cmark & --- & \cmark & --- & 24.9 & --- \\
5 & Hindsight twins & \cmark & \cmark & --- & --- & 29.0 & --- \\
6 & Hindsight twins & \cmark & \cmark & \cmark & --- & 27.5 & --- \\
7 & Cross-author pair & --- & \cmark & --- & \cmark & 29.2 & --- \\
\rowcolor{oursbg}
8 & \textbf{\method (ours)} & \cmark & \cmark & \cmark & \cmark & \best{30.7} & --- \\
\midrule
\multicolumn{8}{@{}l}{\textit{FLUX.2-klein (sunset critic)}} \\
9 & Hindsight twins & --- & --- & --- & --- & 24.0 & 4.7 \\
10 & Hindsight twins & \cmark & --- & --- & --- & \best{25.1} & \best{2.0} \\
\bottomrule
\end{fittab}
\end{table}

\subsection{Ablations}
\label{subsec:ablation}

As shown in Rows 1 to 3 of \cref{tab:ablation}, replacing the positive self-teacher with a contrastive target under a frozen teacher on SD3.5-M raises GM from 21.4 to 25.4 for both the cross-author pair and hindsight twins, while only hindsight twins prevent habit transfer (2.2\% vs.\ 81.9\% Sun.).
Refreshing the teacher each round lifts the unweighted twin from 25.4 to 29.0 GM in Row 5.
Because $w(\sigma)=2\sigma$ doubles the effective step size near $\sigma=1$, applying it without weight averaging increases last-step variance in Rows 4 and 6 (24.9 and 27.5 GM), whereas pairing it with LoRA EMA smooths larger high-noise updates to reach 30.7 GM in Row 8, surpassing the evolving cross-author pair at its best unweighted setting in Row 7 (29.2 GM, where high-noise weighting instead amplifies $\Delta_{\rm drift}$) while an extrapolated positive teacher collapses to 11.4 GM in \cref{sec:supp_results}.
Finally, encoding $r$ and $\rt$ separately without token-state alignment under the sunset critic on FLUX.2-klein lets the critic's scene leak across the shared suffix and lowers GM from 25.1 to 24.0 in Rows 9 and 10, as analyzed in \cref{subsec:local}.

\section{Extended Analysis}
\label{sec:analysis}

This section validates two factors behind \method's effectiveness in student training: (1) canceling rewrite drift in the target while keeping the expanded scene in the template is essential for alignment; and (2) aligning non-slot token states keeps the contrast local, blocking scene leakage and focusing updates on failed bindings. \cref{subsec:noise} further tests robustness to critic errors.

\subsection{Role of the Critic's Scene Rewrite}
\label{subsec:learns}

\cref{subsec:drift} showed in velocity space that rewrite drift $\Delta_{\rm drift}$ dominates whole-prompt updates, whereas contrasting the failed span inside a shared scene isolates the local fix $\Delta_{\rm twin}$. To verify this in student training, we add $\lambda_{\rm d}\,w(\sigma)\,\Delta_{\rm drift}$ from \cref{eq:decomp} back into \method's target on SD3.5-M under a fixed schedule in \cref{tab:dose}. Under the default critic, raising $\lambda_{\rm d}$ from $0$ to $2.0$ lowers GM from 30.7 to 26.5 and J2 from 47.9 to 44.8, cuts plain backgrounds from 76.2\% to 46.8\% at $\lambda_{\rm d}=1.0$, and on one $\lambda_{\rm d}=2.0$ seed amplifies the warm light in 4.5\% of revisions into 25.9\% sunset images; under the sunset critic, $\lambda_{\rm d}=0.5$ already yields 95.4\% sunset images and drops GM from 30.2 to 28.2. 

\begin{table}[t]
\centering
\footnotesize
\renewcommand{\arraystretch}{1.04}
\caption{\textbf{Reintroducing rewrite drift into \method's target} on SD3.5-M (six rounds, evolving teacher). Adding $\lambda_{\rm d}\,w(\sigma)\,\Delta_{\rm drift}$ to \cref{eq:target} ($\lambda=2$, $w(\sigma)=2\sigma$) interpolates from \method ($\lambda_{\rm d}=0$) to the noise-weighted pair ($\lambda_{\rm d}=2.0$). We report EMA GM, J2, and plain-background rate (Plain, \%) under the default critic, and current GM and sunset rate (Sun., \%) under the sunset critic. \textbf{Bold} marks the best trained model.}
\label{tab:dose}
\begin{fittab}{\linewidth}{@{}F{l} ccc cE{c}@{}}
\toprule
& \multicolumn{3}{c}{Default critic} & \multicolumn{2}{c}{Sunset critic} \\
\cmidrule(lr){2-4}\cmidrule(l){5-6}
$\lambda_{\rm d}$ & GM $\uparrow$ & J2 $\uparrow$ & Plain & GM $\uparrow$ & Sun. $\downarrow$ \\
\midrule
\rowcolor{basebg}
Base model & 19.6 & 35.4 & 62.2 & 19.6 & 2.2 \\
\rowcolor{oursbg}
\textbf{0 (\method)} & \best{30.7} & \best{47.9} & 76.2 & \best{30.2} & \best{2.4} \\
0.5 & 30.1 & 47.4 & 63.6 & 28.2 & 95.4 \\
1.0 & 28.9 & 45.9 & 46.8 & 23.6 & 99.5 \\
2.0 (pair) & 26.5 & 44.8 & 53.1\rlap{$^{\ast}$} & 16.0 & 99.6 \\
\bottomrule
\end{fittab}
\end{table}


Restricting the critic from expanding $c$ at the source cannot replace \method either. Tightening the SD3.5-M critic prompt from full-scene revisions (23 median words) to UniEvo~\cite{wu2026unievo} rules (15 words) or emphasis edits (10 words) in Appendix~\cref{tab:critic_style} fails to lift the positive teacher (21.4, 20.1, and 19.5 GM) and still transfers 75.5\% and 99.3\% sunset light to the positive teacher and pair under UniEvo rules. Meanwhile, losing scene context drops \method from 30.9 to 27.2 and 24.8 GM across two seeds as $T$ collapses toward the bare user prompt. Scene expansion must therefore be kept in $T$ for context and canceled in the target.

\subsection{Efficacy of Token-State Alignment}
\label{subsec:local}

Even with a shared template $T$, encoding $r$ and $\rt$ separately lets the slot edit shift surrounding scene tokens via attention in \cref{subsec:align}. Training on identical template pairs $(r,\rt)$ confirms this leakage. On FLUX.2-klein in Rows 9 to 10 of \cref{tab:ablation} and Appendix~\cref{tab:embedding}, independent encoding lowers plain backgrounds from 51.4\% (base) to 45.2\% under the default critic (vs.\ 56.3\% aligned); under the sunset critic, plain backgrounds plunge to 14.9\%, sunset rate rises to 4.7\%, and GM drops from 25.1 to 24.0, whereas suffix alignment keeps 62.3\% plain backgrounds and 2.0\% sunset images. Removing alignment on sunset Qwen-Image-2512 likewise cuts plain backgrounds from 27.1\% to 21.3\% and GM from 49.3 to 47.3.

Fixing non-slot states concentrates updates on failed local bindings: on Qwen-Image-2512 in Appendix~\cref{tab:skills_full}, \method lifts attribute, count, and verb accuracy from 74.6\%, 66.7\%, and 33.9\% to 89.8\%, 74.6\%, and 46.6\% (vs.\ 81.1\% and 35.2\% on attributes and verbs for the pair).

\subsection{Sensitivity to Critic Errors}
\label{subsec:noise}

Because \method relies on critic-diagnosed slots in \cref{subsec:twins}, we test how wrong critiques affect SD3.5-M training by randomly replacing 25\% or 50\% of valid critiques each round with critiques from other failed prompts in \cref{sec:supp_impl}. Across two seeds under EMA weights, \method eases from 30.9 to 30.4 and 28.4 GM and the unweighted pair from 30.2 to 29.7 and 27.8, with habit rates unchanged: unlike systematic rewrite drift, mismatched contrasts are uncorrelated across prompts and damped by the anchor $\vel_{\bar\theta}(c)$ in \cref{eq:target}, so \method keeps its lead.

\section{Related Work}
\label{sec:related}

\paragraph{Post-training text-to-image models with feedback.}
While reward models and vision-language critics can readily assess generated images, turning their feedback into stronger single-pass generation remains challenging. Existing approaches fall into three categories: test-time revision, scalar-reward reinforcement learning, and multimodal self-rewarding loops. Test-time prompt rewriting~\cite{hao2023promptist} and iterative self-reflection~\cite{madaan2023selfrefine,zhuo2025reflection} refine generations without updating model weights, but add extra VLM calls and resampling to every user request. Post-training methods instead internalize feedback into the generator so inference stays single-pass: preference optimization and online policy-gradient methods such as DDPO~\cite{black2024ddpo}, Diffusion-DPO~\cite{wallace2024diffusiondpo}, Flow-GRPO~\cite{liu2025flowgrpo}, DanceGRPO~\cite{xue2025dancegrpo}, and DiffusionNFT~\cite{zheng2025diffusionnft} optimize generators from external reward models, though group-based policy gradients require multiple rollouts per prompt to estimate relative advantages reliably. Unified self-improvement frameworks including SRUM~\cite{jin2025srum}, UniCorn~\cite{han2026unicorn}, and Dreaming in Flow~\cite{hao2026dreaming} remove separate reward models by critiquing self-generated images with a VLM, yet still convert feedback into scalar rewards or sample-level supervision rather than dense trajectory-level corrections. Unlike test-time revision that slows inference or reward-based post-training that discards localized error diagnoses and needs multiple rollouts per prompt, our proposed \method turns structured VLM critiques from a single rollout per prompt into dense trajectory supervision, enabling self-improvement while keeping inference single-pass and VLM-free.

\paragraph{On-policy self-distillation.}
To obtain dense step-wise supervision from a single rollout per prompt~\cite{agarwal2024gkd,lu2025opdblog,li2026video}, on-policy distillation trains a student along its own trajectories. Extending this paradigm to diffusion and flow-matching models, recent methods query a stronger external teacher~\cite{li2026diffusionopd,fang2026flowopd}, extrapolate along policy or guidance differences under a shared prompt~\cite{yang2026gopd,wei2026stepopd,xu2026sg,lin2026dreopd,li2026rethinkingcfg}, or perform teacher-free self-distillation~\cite{zhou2026diffusionopsd,zhang2026selfopd}. Following privileged-context self-distillation in language models~\cite{vapnik2009lupi,snell2022context,zhao2026opsd,hubotter2026sdpo,pan2026rlcsd,baumann2026repulsive,pei2026nsd}, visual self-teachers condition the generator itself on reference images~\cite{jiang2026dopsd,bu2026hpsd} or critic-rewritten prompts~\cite{lin2026peopsd,wu2026unievo,wang2026dopcd}. Distilling an expanded prompt directly works well when the goal is to internalize general prompt enhancement, as in PE-OPSD~\cite{lin2026peopsd}. When correcting generation errors, however, the user's prompt already states the desired requirement, so replacing only the failed span leaves it unchanged and forces the critic to expand and rewrite the entire scene; distilling this full revision~\cite{wu2026unievo,wang2026dopcd} entangles the local fix with unrequested scenery that outweighs the correction, and refreshing an evolving teacher~\cite{tarvainen2017meanteacher} each round compounds those critic habits much as iterative self-training amplifies distribution shifts~\cite{alemohammad2024mad,shumailov2024collapse}. Our proposed \method resolves these limitations by pairing the intended fix with a hindsight twin of the observed error inside a shared slotted critic template, aligning non-slot token states in feature space so the slot edit does not leak into surrounding scene tokens, and focusing the contrast on high-noise steps~\cite{jeong2025diffusionclassifiers} before refreshing the teacher.\nocite{kaushik2020cad,gandikota2024sliders,meyarian2026direct,hertz2023p2p}

\section{Conclusion}
\label{sec:conclusion}

Text-to-image self-distillation inherits a critic's stylistic habits because useful revisions expand the whole scene, leaving updates dominated by scene rewrite rather than the fix. We propose Hindsight Twins (\method), which pairs the intended fix and observed error inside a slotted critic template and aligns non-slot token states in feature space, compounding only the span-local correction under an evolving teacher. Across three rectified-flow backbones, \method improves compositional alignment and zero-shot transfer, keeps habit rates at base levels under biased critics, and sustains its gains over twelve rounds where contrastive baselines decay.


{
    \small
    \linespread{0.97}\selectfont
    \setlength{\bibsep}{0pt}
    \bibliographystyle{ieeenat_fullname}
    \bibliography{main}
}

\clearpage
\maketitlesupplementary
\appendix
\setlength{\aboverulesep}{0.25ex}
\setlength{\belowrulesep}{0.40ex}
\renewcommand{\theHsection}{supp.\Alph{section}}
\renewcommand{\dbltopfraction}{0.94}
\renewcommand{\dblfloatpagefraction}{0.82}
\renewcommand{\topfraction}{0.92}
\renewcommand{\floatpagefraction}{0.82}
\renewcommand{\textfraction}{0.05}
\setcounter{dbltopnumber}{4}
\setcounter{topnumber}{4}
\setcounter{totalnumber}{6}
\captionsetup[table]{skip=2.0pt}
\setlength{\dblfloatsep}{3.2pt plus 1pt minus 1pt}
\setlength{\dbltextfloatsep}{3.8pt plus 1pt minus 1pt}
\makeatletter
\setlength{\@fptop}{0pt}
\setlength{\@fpsep}{6pt}
\setlength{\@fpbot}{0pt plus 1fil}
\setlength{\@dblfptop}{0pt}
\setlength{\@dblfpsep}{6pt}
\setlength{\@dblfpbot}{0pt plus 1fil}
\makeatother

\cref{sec:supp_impl} gives training hyperparameters, baseline details, and wall-clock costs. \cref{sec:supp_critic} describes the critic prompt, twin filtering, alternative critic models, and velocity probes across backbones. \cref{sec:supp_eval} covers evaluation judges, sampling variance, and habit-probe validation; \cref{sec:supp_results} reports extended benchmark, multi-round, and ablation results; and \cref{sec:supp_lim} discusses limitations and future work.

\section{Implementation Details}
\label{sec:supp_impl}

\subsection{Training Setup and Compute Cost}

\begin{table*}[t]
\centering
\footnotesize
\renewcommand{\arraystretch}{1.08}
\caption{\textbf{Training and evaluation settings across backbones.} Wall-clock times are for \method excluding evaluation; Qwen-Image-2512 lists the 400-prompt held-out split alongside the 800-prompt settings (shared by the disjoint and full 800-prompt protocols; see below).}
\label{tab:supp_hparams}
\begin{fittab}{\textwidth}{@{}F{l} cc cE{c}@{}}
\toprule
& \multicolumn{2}{c}{Qwen-Image-2512} & & \\
\cmidrule(lr){2-3}
Setting & Held-out (400) & 800-prompt & SD3.5-M & FLUX.2-klein \\
\midrule
Resolution, sampling steps, and CFG & \multicolumn{2}{c}{$1024^2$, 20, 4.0} & $512^2$, 28, 4.5 & $512^2$, 28, 4.0 \\
Training / evaluation prompts & 400 / 400 & 800 / 800 & 400 / 400 & 400 / 400 \\
Rounds $\times$ updates per round & \multicolumn{2}{c}{$4\times60$} & $6\times60$ & $6\times60$ \\
LoRA rank / $\alpha$ (trainable params) & \multicolumn{2}{c}{16 / 32 (47.2M)} & 32 / 64 (18.8M) & 32 / 64 (37.4M) \\
Global batch size / learning rate / EMA & \multicolumn{2}{c}{16 / $10^{-4}$ / 0.98} & 64 / $10^{-4}$ / 0.98 & 64 / $10^{-4}$ / 0.98 \\
\rowcolor{oursbg}
\textbf{\method teacher, $\lambda$, and alignment} & \multicolumn{2}{c}{\cellcolor{oursbg}Evolving, 4, suffix} & Evolving, 2, span & Hybrid, 1, suffix \\
Evaluation images & $400\times4$ & $800\times1$ & $400\times4$ & $400\times4$ \\
Per-round time (sample + critique / updates) & 10.9 / 6.4 min & 19.5 / 5.5 min & 2.2 / 1.0 min & 3.5 / 2.0 min \\
Total training time & 69 min & 100 min & 19 min & 33 min \\
\bottomrule
\end{fittab}
\end{table*}

\begin{algorithm*}[t]
\caption{Hindsight Twins (\method): span-local self-distillation with an evolving teacher}
\label{alg:hint}
\small
\begin{algorithmic}[1]
\Require student $\theta$, frozen base $\theta_0$, prompts $\mathcal P$, critic $\mathcal C$, rounds $K$, updates $S$, contrast scale $\lambda$
\State Initialize the student LoRA exponential moving average $\theta_{\mathrm{EMA}}\gets\theta$
\For{$k=1,\dots,K$}
  \State $\bar\theta\gets\theta$; $\theta_\Delta\gets\bar\theta$ \Comment{use $\theta_\Delta=\theta_0$ for the hybrid teacher on FLUX.2-klein}
  \State Sample one rollout $x_0\sim p_\theta(\cdot\mid c)$ for each $c\in\mathcal P$ and critique it with $\mathcal C$
  \State Retain accepted rollouts and valid failures; drop critiques that fail slot checks
  \State For each valid failure, form $r=T(f)$ and $\rt=T(a)$, and copy $r$'s token states onto $\rt$ outside the slot
  \For{$s=1,\dots,S$}
    \State Sample a minibatch (with 10\% unconditional prompts $c=\varnothing$), $\sigma\sim\mathcal U(0.05,0.95)$, and $\epsilon\sim\mathcal N(0,I)$
    \State Set $x_\sigma=(1-\sigma)x_0+\sigma\epsilon$ and initialize the anchor $\vel^\star\gets\vel_{\bar\theta}(x_\sigma,\sigma,c)$
    \State On failed rollouts, add $2\lambda\sigma[\vel_{\theta_\Delta}(x_\sigma,\sigma,r)-\vel_{\theta_\Delta}(x_\sigma,\sigma,\rt)]$ to $\vel^\star$
    \State Update $\theta$ on $\|\vel_\theta(x_\sigma,\sigma,c)-\mathrm{sg}[\vel^\star]\|^2$ and update $\theta_{\mathrm{EMA}}$
  \EndFor
\EndFor
\end{algorithmic}
\end{algorithm*}

\paragraph{Hyperparameters and evaluation seeds.} Within each backbone, every method shares the same prompts, rollout budget, critic, update schedule, LoRA rank, and evaluation split listed in \cref{tab:supp_hparams}. On Qwen-Image-2512 and SD3.5-M, \method evaluates both the anchor $\vel_{\bar\theta}(c)$ and the twin contrast with the round-start teacher $\bar\theta$, using $\lambda=2$ on SD3.5-M and $\lambda=4$ on Qwen-Image-2512 to match its higher-dimensional $1024^2$ latent space and smaller relative twin norms in \cref{tab:probe_qwen}. On FLUX.2-klein, \method uses a hybrid teacher in \cref{alg:hint} that refreshes the anchor $\vel_{\bar\theta}(c)$ from the student each round while evaluating $\vel_{\theta_0}(r)-\vel_{\theta_0}(\rt)$ with the frozen base $\theta_0$ ($\lambda=1$) to keep slot sensitivity stable across rounds. Main tables average three seeds on Qwen-Image-2512 and in \cref{tab:habit,tab:skills_full}, four seeds for \method and the cross-author pair on SD3.5-M and FLUX.2-klein, and two to three seeds for the remaining baselines; all methods reuse the fixed settings of \cref{tab:supp_hparams} without retuning on the 800-prompt disjoint benchmark in \cref{tab:qwen}, the habit stress tests in \cref{tab:habit}, or the zero-shot benchmarks in \cref{tab:ood}.

\paragraph{Token-state alignment across encoder architectures.} Expanding \cref{subsec:align}, we encode $r=T(f)$ and $\rt=T(a)$ separately through each backbone's text encoder and copy $r$'s hidden states onto $\rt$ outside the edited slot. Under bidirectional encoders, the slot edit propagates to both sides of the prompt: on SD3.5-M, which combines CLIP-L and CLIP-G token features with bidirectional T5-XXL states, \emph{span replacement} copies $r$'s hidden states onto $\rt$ at all positions outside the differing token span in each encoder while retaining $\rt$'s pooled CLIP projections. Under causal encoders, namely Qwen2.5-VL in Qwen-Image-2512 and Qwen3 in FLUX.2-klein, tokens before the slot cannot attend to it and already share identical states, so alignment reduces to \emph{suffix alignment}. Specifically, we match the longest common prefix and valid suffix of the two unpadded token sequences (accommodating slots of different token lengths) and copy $r$'s hidden states onto the corresponding valid suffix positions of $\rt$, leaving slot and padding tokens untouched. Following each backbone's conditioning format, this step operates on Qwen2.5-VL's last-layer states (after its 34-token system prefix) in Qwen-Image-2512 and on stacked Qwen3 layers 9, 18, and 27 (padded to 512 tokens) in FLUX.2-klein.

\paragraph{Training batches and anchor samples.} In each round (\cref{alg:hint}), the critic splits rollouts into accepted and failed samples, dropping any failure whose critique fails the slot checks in \cref{sec:supp_critic}. \method trains on both groups: failed samples use the target in \cref{eq:target}, while accepted samples use target $\vel_{\bar\theta}(c)$ so the student preserves behaviors that already work. Following UniEvo-VL~\cite{wu2026unievo}, the positive self-teacher trains only on failed samples; adding accepted samples with target $\vel_{\bar\theta}(c)$ slightly lowers its SD3.5-M score from 21.4 to 20.8 GM (vs.\ Row 1 of \cref{tab:ablation}), confirming that \method's gains come from the twin target rather than from keeping accepted samples. Finally, across all distillation methods, each minibatch includes 10\% empty-prompt samples ($c=\varnothing$) with target $\vel_{\bar\theta}(\varnothing)$ so that the unconditional prediction used in classifier-free guidance does not drift.

\paragraph{Training time and compute cost.} \cref{tab:supp_hparams} reports per-round and total wall-clock times across backbones (19 to 100 minutes in total; about 39 and 67 minutes for 12-round extensions on SD3.5-M and FLUX.2-klein). Every method critiques each rollout once per round, and on a failed sample \method runs three teacher forward passes ($c$, $r$, and $\rt$), matching the cross-author pair ($c$, $r$, and $h$) and adding one pass over the positive self-teacher ($c$ and $r$). Because rollout sampling and VLM critique take 65\% to 78\% of each round, this extra pass adds under 4\% end-to-end training time: across four 800-prompt rounds on Qwen-Image-2512, \method takes 99.7 minutes, matching the cross-author pair (99.8 minutes) and staying within 5 minutes of the positive self-teacher (96.4 minutes) and single-rollout DiffusionNFT (94.9 minutes). By comparison, four-rollout DiffusionNFT takes 361.5 minutes, and sample-level SFT takes 80.4 minutes on the 400-prompt split (vs.\ 69.2 minutes for \method) because failed prompts must be regenerated and re-verified. At inference time, the LoRA adapter merges into the base weights with zero extra latency or VLM calls.

\subsection{Baseline and Analysis Protocols}

\paragraph{Self-distillation baselines.} For fair comparison in \cref{tab:heldout,tab:heldout_full}, each distillation baseline is tuned over its own teacher schedule (frozen $\theta_0$, step-wise EMA~\cite{tarvainen2017meanteacher}, or round-start evolving), student LoRA EMA, noise weighting, and contrast scale $\lambda$, and we report its strongest setting. The positive self-teacher uses a step-wise EMA teacher on Qwen-Image-2512 and a frozen base teacher $\theta_0$ without student LoRA EMA on SD3.5-M and FLUX.2-klein, since refreshing it compounds rewrite drift. Sharing \method's accepted-plus-failed batch split, the cross-author pair uses unweighted updates ($w(\sigma)=1$, avoiding high-noise amplification of $\Delta_{\rm drift}$) with an evolving teacher on Qwen-Image-2512 and FLUX.2-klein and a step-wise EMA teacher on SD3.5-M, whereas user-worded hindsight contrast uses the evolving schedule and is omitted from FLUX.2-klein zero-shot and twelve-round evaluations after collapsing on held-out prompts (\cref{tab:heldout}).

\paragraph{RL and sample-level SFT baselines.} DiffusionNFT~\cite{zheng2025diffusionnft} runs inside the same training loop under an identical rollout budget and critic, using the fraction of satisfied GenEval2 questions per rollout as the scalar reward and the round-start student as the old policy at a tuned learning rate ($10^{-4}$, $3\times10^{-4}$, and $3\times10^{-5}$ on Qwen-Image-2512, SD3.5-M, and FLUX.2-klein); rewards are normalized across the round when using a single rollout per prompt, and within each prompt group following~\cite{zheng2025diffusionnft} at four rollouts per prompt (\cref{tab:qwen_full}). For SFT on the revised prompt, each failed rollout is regenerated from $r$ with the same initial noise and kept only if the critic verifies the new image against the original prompt $c$ (15\%, 7\%, and 9\% pass rates); accepted regenerations are relabeled with $c$ and pooled with initially accepted samples for standard flow-matching fine-tuning without student EMA at a tuned learning rate ($10^{-4}$, $10^{-5}$, and $3\times10^{-5}$; three seeds on Qwen-Image-2512 and single runs on SD3.5-M and FLUX.2-klein).

\paragraph{Two 800-prompt protocols on Qwen-Image-2512.} Prior self-evolution work on GenEval2~\cite{wu2026unievo} treats the benchmark's 800 prompts as an unlabeled prompt set, training across rounds directly on all 800 prompts via VLM critiques (without reference images) and evaluating on the same 800 prompts at a fixed evaluation seed ($800\times1$ in \cref{tab:supp_hparams}). We refer to this in-benchmark setup as the \emph{full 800-prompt protocol} (\cref{tab:habit,tab:rounds,tab:qwen_full,tab:skills_full}). Because training on the test prompts themselves can conflate generalizable binding with prompt-specific fitting, \cref{tab:qwen} (and the top block of \cref{tab:qwen_full}) introduces a stricter \emph{disjoint-prompt protocol}: we re-implemented GenEval2's compositional grammar over objects, attributes, counts, positions, and verbs~\cite{kamath2025geneval2} together with its question generator, verified benchmark coverage, and sampled 800 new training prompts (100 per complexity level from 3 to 10 atoms) with zero string overlap against GenEval2, training all methods under the same \cref{tab:supp_hparams} settings before evaluating on the 800 GenEval2 benchmark prompts.

\paragraph{Reintroducing rewrite drift and simulated critic errors.} \cref{tab:dose_full,tab:noise_full} give the full results for \cref{sec:analysis}. In \cref{tab:dose,tab:dose_full}, we add rewrite drift back into \method's failed-sample target (\cref{eq:target}), setting $\vel^\star=\vel_{\bar\theta}(c)+w(\sigma)[\lambda\,\Delta_{\rm twin}+\lambda_{\rm d}\,\Delta_{\rm drift}]$ while keeping all other settings fixed. Since $\vel_{\bar\theta}(r)-\vel_{\bar\theta}(h)=\Delta_{\rm twin}+\Delta_{\rm drift}$ (\cref{eq:decomp}), $\lambda_{\rm d}=0$ is \method and $\lambda_{\rm d}=\lambda=2.0$ becomes the noise-weighted cross-author pair. In \cref{subsec:noise,tab:noise_full}, we test sensitivity to wrong critiques by randomly selecting 25\% or 50\% of valid failed samples in each round and replacing each sample's critique with one from another failed prompt in that round, using the same random seeds across methods.

\section{Critic and Twin Construction}
\label{sec:supp_critic}

\subsection{Critic Prompt, Twin Filtering, and Critic Choice}

\paragraph{Critic prompt and output format.} For each rollout, the critic receives the generated image, the user prompt $c$, and its verification questions (GenEval2 atomic questions, or questions derived from GenEval metadata). Using the prompt below, it returns a JSON object containing pass/fail flags for each question and, when any requirement fails, a list of violations (\texttt{span} $s_i$, \texttt{actual} $a_i$, and \texttt{fix} $f_i$) together with a slotted scene rewrite $T$ (\texttt{revised\_template}). In the habit stress tests (\cref{subsec:habit}), we append the single \emph{Habit suffix} line shown at the bottom of the box, leaving the verification questions and violation fields unchanged.

\begin{figure}[!t]
\begin{tcolorbox}[enhanced, colback=gray!4, colframe=gray!45, boxrule=0.4pt, arc=1.5pt, left=1.5mm, right=1.5mm,
  top=1.0mm, bottom=1.0mm, before skip=0pt, after skip=0pt, fontupper=\footnotesize]
\textbf{System:} You are a strict, precise verifier of text-to-image generations. Answer only with JSON following the schema.\\[2pt]
\textbf{User:} Prompt given to the image generator: \texttt{"\{prompt\}"}\\
Check the image against each atomic requirement. For each, write what the image actually shows (\texttt{observed}) and whether it matches the expected answer (\texttt{ok}): \texttt{\{atoms\}}.\\
If any requirement fails, fill \texttt{violations}, one entry per failed aspect of the prompt:
\begin{itemize}[leftmargin=1.1em, itemsep=0.5pt, topsep=1pt]
  \item \texttt{span}: exact substring of the prompt violated by the image (\eg, \texttt{"three apples"}, \texttt{"red car"}).
  \item \texttt{actual}: natural phrase in the same grammatical role describing what the image shows instead (\eg, \texttt{"two apples"}, \texttt{"no <object>"}), with no explanations or parentheses.
  \item \texttt{fix}: the phrase as required in a correct prompt (normally identical to \texttt{span}; \texttt{fix} $\neq$ \texttt{actual}).
\end{itemize}
Then write \texttt{revised\_template}: an improved, more explicit version of the prompt, in the style of an expert prompt engineer, that restates the whole scene in new words and stresses the failed requirements; put \texttt{\{0\}}, \texttt{\{1\}}, \dots\ exactly once each where each violation's phrase goes. Surrounding words must read naturally with both \texttt{fix} and \texttt{actual}, without repeating properties carried inside the slots.\\[2pt]
\textbf{Habit suffix:} \texttt{Style rule for revised\_template: always describe the scene as [lit by warm, golden sunset light | lit by cold, blue, overcast light | a vintage film photograph with faded colors and film grain].}
\end{tcolorbox}
\vspace{-4pt}
\end{figure}

\paragraph{Why revisions expand the prompt.} Across all 1,341 valid first-round violations on Qwen-Image-2512 (377), SD3.5-M (464), and FLUX.2-klein (500), the critic's intended fix $f_i$ is identical to the original span $s_i$ in $c$ (\eg, ``seven black cows''). Simply substituting $f_i$ back into $c$ therefore leaves the prompt unchanged and gives zero update ($\Delta_{\rm pos}=0$). Prior self-teachers ask the critic to expand the prompt for this exact reason: UniEvo-VL~\cite{wu2026unievo} requests ``one complete, self-contained new T2I prompt'' that is ``never a terser paraphrase,'' PE-OPSD~\cite{lin2026peopsd} fills in ``otherwise unspecified attributes, composition, and scene details,'' and D-OPCD~\cite{wang2026dopcd} distills agent-rewritten prompts. Under our default Qwen3-VL-32B critic, user prompts have a median length of 10 words, whereas revised prompts expand to 25, 24, and 22 median words on Qwen-Image-2512, SD3.5-M, and FLUX.2-klein, adding background or lighting terms absent from $c$ in 74\%, 61\%, and 59\% of revisions.

\paragraph{Slot checks.} Before building $(r,\rt)$ from a critique, we apply two simple string cleanups: matching leading articles (``a'', ``an'', ``the'') to $c$, and removing words right after \texttt{\{i\}} if they already appear at the end of both $f_i$ and $a_i$ (\eg, turning ``\texttt{\{0\}} a sandwich'' into ``\texttt{\{0\}}'' when the slot phrase is ``right of a sandwich''). We then accept the critique if each failed span $s_i$ appears once in $c$, spans do not overlap, $f_i\neq a_i$, each placeholder \texttt{\{i\}} appears once in $T$, and neither $r=T(f)$ nor $\rt=T(a)$ contains repeated adjacent words. Any failed rollout whose critique fails these checks is skipped for that round (\cref{alg:hint}), with no extra VLM retries.

\paragraph{Critic model choice.} Jointly diagnosing visual errors and writing a clean slotted template requires reliable instruction following, so we use a 30B-scale open-source VLM as our default critic. In a single pass, Qwen3-VL-32B produces valid twins on 78\% of first-round failures on Qwen-Image-2512, 69\% to 73\% across SD3.5-M runs, and 77\% on FLUX.2-klein. Swapping Qwen3-VL-32B for InternVL3.5-38B~\cite{wang2025internvl35} on SD3.5-M without changing the prompt still raises EMA GM from 19.6 (base) and 20.1 (positive self-teacher) to 25.4 with \method (vs.\ 25.1 for the cross-author pair), showing that \method is not tied to a single VLM family.

\begin{table*}[t]
\centering
\footnotesize
\renewcommand{\arraystretch}{1.08}
\caption{\textbf{Restricting critic rewrite length on SD3.5-M.} Across two seeds (mean $\pm$ std; setup in \cref{sec:supp_critic}), we compare full-scene rewrites (\emph{Full}), shorter rewrites following UniEvo-VL~\cite{wu2026unievo} (\emph{UniEvo}), and emphasis-only edits (\emph{Minimal}, omitted under the sunset critic as it forbids lighting additions). Words and Valid give median revision length and slot-check pass rate on base failures; \textbf{bold} marks the best trained GM and Sun.\ rate per critic.}
\label{tab:critic_style}
\begin{fittab}{\textwidth}{@{}F{l} cc l ccE{c}@{}}
\toprule
Revision style & Words & Valid (\%) & Method & GM $\uparrow$ & Plain (\%) & Sun. (\%) $\downarrow$ \\
\midrule
\rowcolor{basebg}
Base model & --- & --- & --- & 19.6\sdz{0.0} & 62.2\sdz{0.0} & \phantom{0}2.2\sdz{0.0} \\
\midrule
\multicolumn{7}{@{}l}{\textit{Default critic}} \\
Full (default) & 23 & 73 & Positive teacher & 21.4\sd{0.4} & 48.2\sd{6.3} & \phantom{0}3.2\sd{0.5} \\
 & & & Cross-author pair & 30.2\sd{0.8} & 54.3\sd{4.8} & \phantom{0}3.3\sd{0.1} \\
\rowcolor{oursbg}
\cellcolor{white} & \cellcolor{white} & \cellcolor{white} & \textbf{\method (ours)} & \bestpm{30.9}{0.5} & 73.8\sd{0.7} & \phantom{0}\bestpm{1.7}{0.2} \\
\cmidrule(l){1-7}
UniEvo~\cite{wu2026unievo} & 15 & 56 & Positive teacher & 20.1\sd{0.3} & 61.8\sd{0.7} & \phantom{0}2.3\sd{0.3} \\
 & & & Cross-author pair & 26.6\sd{1.1} & 68.6\sd{1.3} & \phantom{0}\bestpm{1.7}{0.3} \\
\rowcolor{oursbg}
\cellcolor{white} & \cellcolor{white} & \cellcolor{white} & \textbf{\method (ours)} & 27.2\sd{1.0} & 81.2\sd{1.8} & \phantom{0}\bestpm{1.7}{0.0} \\
\cmidrule(l){1-7}
Minimal (emphasis only) & 10 & 64 & Positive teacher & 19.5\sd{0.3} & 60.7\sd{0.8} & \phantom{0}2.3\sd{0.1} \\
 & & & Cross-author pair & 24.9\sd{0.2} & 68.3\sd{4.7} & \phantom{0}1.8\sd{0.1} \\
\rowcolor{oursbg}
\cellcolor{white} & \cellcolor{white} & \cellcolor{white} & \textbf{\method (ours)} & 24.8\sd{1.9} & 74.6\sd{11.2} & \phantom{0}2.8\sd{1.3} \\
\midrule
\multicolumn{7}{@{}l}{\textit{Sunset critic}} \\
Full (default) & 31 & 76 & Positive teacher & 19.4\sd{1.0} & 11.5\sd{2.2} & 83.0\sd{2.8} \\
 & & & Cross-author pair & 16.9\sd{2.3} & \phantom{0}6.0\sd{1.3} & 99.5\sd{0.0} \\
\rowcolor{oursbg}
\cellcolor{white} & \cellcolor{white} & \cellcolor{white} & \textbf{\method (ours)} & \bestpm{30.3}{0.6} & 75.2\sd{6.3} & \phantom{0}\bestpm{2.3}{0.1} \\
\cmidrule(l){1-7}
UniEvo~\cite{wu2026unievo} & 33 & 53 & Positive teacher & 18.8\sd{0.8} & 19.3\sd{4.4} & 75.5\sd{6.3} \\
 & & & Cross-author pair & 16.0\sd{0.5} & 15.6\sd{8.3} & 99.3\sd{0.7} \\
\rowcolor{oursbg}
\cellcolor{white} & \cellcolor{white} & \cellcolor{white} & \textbf{\method (ours)} & 23.7\sd{1.8} & 65.1\sd{0.1} & \phantom{0}4.4\sd{2.5} \\
\bottomrule
\end{fittab}
\end{table*}

\begin{table}[t]
\centering
\footnotesize
\renewcommand{\arraystretch}{1.10}
\caption{\textbf{Alignment gains by critic rewrite share.} Held-out prompts (\cref{tab:heldout}, EMA weights) are split into three equal groups by the fraction of revision words absent from $c$ on base-model outputs; rows report base GM and paired GM gains ($\pm$ standard error across prompts).}
\label{tab:rewrite_buckets}
\begin{fittab}{\linewidth}{@{}F{l} c c E{c}@{}}
\toprule
& \multicolumn{3}{c@{}}{Prompt group by critic rewrite share} \\
\cmidrule(l){2-4}
Metric & Low & Mid & High \\
\midrule
\multicolumn{4}{@{}l}{\textit{Qwen-Image-2512} {\scriptsize\textcolor{sdgray}{(346 prompts)}}} \\
Rewrite share (\%) & 0 to 45 & 45 to 60 & 60 to 89 \\
\rowcolor{basebg}
Base model GM & 25.7 & 27.3 & 30.6 \\
\method vs.\ Positive teacher & \phantom{0}${+}$8.5\sd{1.7} & ${+}$11.5\sd{1.9} & ${+}$13.6\sd{2.0} \\
\method vs.\ Cross-author pair & \phantom{0}${+}$0.8\sd{1.7} & \phantom{0}${+}$2.4\sd{1.8} & \phantom{0}${+}$4.9\sd{2.1} \\
\rowcolor{oursbg}
\textbf{\method vs.\ Base model} & ${+}$10.4\sd{1.8} & ${+}$12.8\sd{1.9} & ${+}$15.5\sd{2.3} \\
\midrule
\multicolumn{4}{@{}l}{\textit{SD3.5-M} {\scriptsize\textcolor{sdgray}{(368 prompts)}}} \\
Rewrite share (\%) & 0 to 40 & 40 to 57 & 57 to 88 \\
\rowcolor{basebg}
Base model GM & 15.0 & 13.5 & 21.0 \\
\method vs.\ Positive teacher & \phantom{0}${+}$6.4\sd{1.3} & \phantom{0}${+}$7.1\sd{1.4} & ${+}$15.1\sd{2.0} \\
\method vs.\ Cross-author pair & \phantom{0}${+}$0.8\sd{1.0} & \phantom{0}${-}$0.1\sd{0.9} & \phantom{0}${+}$1.8\sd{1.4} \\
\rowcolor{oursbg}
\textbf{\method vs.\ Base model} & \phantom{0}${+}$8.2\sd{1.4} & \phantom{0}${+}$9.8\sd{1.5} & ${+}$17.7\sd{1.9} \\
\midrule
\multicolumn{4}{@{}l}{\textit{FLUX.2-klein} {\scriptsize\textcolor{sdgray}{(362 prompts)}}} \\
Rewrite share (\%) & 0 to 38 & 38 to 55 & 55 to 88 \\
\rowcolor{basebg}
Base model GM & 17.8 & 17.2 & 21.5 \\
\method vs.\ Positive teacher & \phantom{0}${+}$5.5\sd{1.1} & \phantom{0}${+}$5.9\sd{1.0} & \phantom{0}${+}$6.7\sd{1.2} \\
\method vs.\ Cross-author pair & \phantom{0}${-}$1.4\sd{0.9} & \phantom{0}${-}$2.0\sd{1.1} & \phantom{0}${-}$1.5\sd{1.4} \\
\rowcolor{oursbg}
\textbf{\method vs.\ Base model} & \phantom{0}${+}$3.8\sd{1.1} & \phantom{0}${+}$2.6\sd{1.1} & \phantom{0}${+}$2.3\sd{1.4} \\
\bottomrule
\end{fittab}
\end{table}

\subsection{Sensitivity to Critic Rewrite Length}

\paragraph{Restricting the critic's rewrite.} \cref{tab:critic_style} tests whether restricting the critic to shorter revisions can prevent rewrite drift on SD3.5-M without \method. Alongside our default full-scene rewrite (23 median words), we evaluate two stricter critic prompts across two seeds: one adapted from UniEvo-VL's synthesis rules, which asks the critic not to add details beyond its diagnostic notes (15 median words), and an emphasis-only prompt that keeps $c$ intact except for emphasis words such as \emph{clearly}, \emph{distinct}, or \emph{exact} (10 median words). Full-scene rewrites lift the positive self-teacher above the 19.6 base model to 21.4 GM, though rewrite drift limits this gain and cuts plain backgrounds from 62.2\% to 48.2\%. Restricting the rewrite fails to help the positive self-teacher: its score drops to 20.1 GM under UniEvo rules and back to the base level (19.5 GM) under minimal edits where $r\approx c$ gives almost no training signal, while under the sunset critic UniEvo rules still transfer sunset lighting to 75.5\% of positive-teacher images and 99.3\% of cross-author pair images. Removing scene context from $T$ also lowers alignment across all methods, dropping \method from 30.9 to 27.2 and 24.8 GM and the cross-author pair from 30.2 to 26.6 and 24.9 GM. Even under these stricter prompts, \method remains the best-performing method overall: it leads both baselines under Full and UniEvo on the default (30.9 and 27.2 GM) and sunset critics (30.3 and 23.7 GM, with only 2.3\% and 4.4\% sunset images), and under minimal edits where $\rt\approx h$ it matches the cross-author pair in GM (24.8 vs.\ 24.9, over 5 points above the positive teacher) with fewer background additions (74.6\% vs.\ 68.3\% plain). Consistent with \cref{subsec:learns}, scene expansion should therefore be kept in $T$ for context and canceled in the target rather than restricted in the critic prompt.

\paragraph{Grouping prompts by rewrite amount.} To check whether heavy scene rewriting hurts \method, we ran the default critic on the 1,600 base-model held-out images per backbone, computed each prompt's rewrite share (the fraction of revision words not in $c$), and split the prompts into three equal groups in \cref{tab:rewrite_buckets}. Shorter user prompts leave more room for the critic to add detail and are also easier for the base model (\eg, 21.0 vs.\ 15.0 base GM on SD3.5-M), so we compare methods on the same prompts within each group. Across all three backbones, the more the critic rewrites a prompt, the further \method pulls ahead of the positive self-teacher (from $+8.5$ to $+13.6$ GM on Qwen-Image-2512, and from $+6.4$ to $+15.1$ on SD3.5-M). Against the cross-author pair, \method's lead grows from $+0.8$ to $+4.9$ GM on Qwen-Image-2512 and holds steady across groups on SD3.5-M and FLUX.2-klein, showing that heavy scene rewriting does not hurt \method.

\subsection{Velocity Probes Across Backbones}

\begin{figure*}[t]
  \centering
  \includegraphics[width=\linewidth]{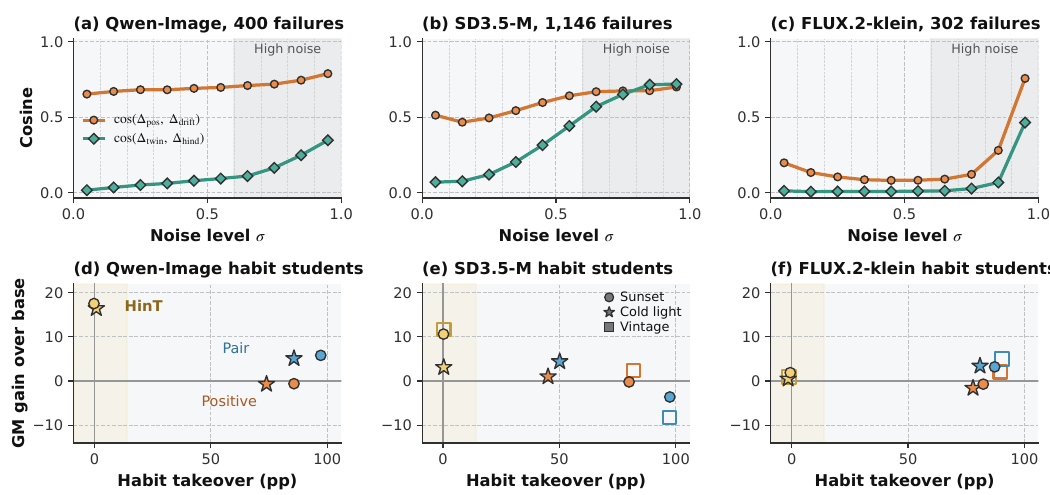}
  \caption{\textbf{Velocity probes and habit transfer across all three backbones.} Top (a to c): median cosines across noise levels (shaded: $\sigma\ge0.6$) from \cref{eq:decomp}; at high noise on every backbone, the full-revision update follows rewrite drift ($\cos(\Delta_{\rm pos},\Delta_{\rm drift})$) while the two local fixes agree ($\cos(\Delta_{\rm twin},\Delta_{\rm hind})$). Bottom (d to f): Soft-TIFA GM gain over the base model versus habit takeover (probe rate minus base rate, in percentage points) across all habit-critic runs in \cref{tab:habit}.}
  \label{fig:motivation_supp}
\end{figure*}

\begin{table}[t]
\centering
\footnotesize
\renewcommand{\arraystretch}{1.08}
\caption{\textbf{Velocity probe on Qwen-Image-2512.} Each row reports the median across 400 first-round base-model failures at noise level $\sigma$: directional cosines between $\Delta_{\rm pos}$ and $\Delta_{\rm drift}$ ($\mathrm{p}\cdot\mathrm{d}$), $\Delta_{\rm twin}$ and $\Delta_{\rm hind}$ ($\mathrm{t}\cdot\mathrm{h}$), and $\Delta_{\rm pos}$ and $\Delta_{\rm twin}$ ($\mathrm{p}\cdot\mathrm{t}$), the share of $\Delta_{\rm pos}$ energy along $\Delta_{\rm twin}$ (Share), and squared $L_2$ norms $E=\|\Delta\|_2^2$.}
\label{tab:probe_qwen}
\begin{fittab}{\linewidth}{@{}F{c} ccc c cc E{c}@{}}
\toprule
& \multicolumn{3}{c}{Directional cosine} & & \multicolumn{3}{c@{}}{Squared norm} \\
\cmidrule(lr){2-4}\cmidrule(l){6-8}
$\sigma$ & $\mathrm{p}\cdot\mathrm{d}$ & $\mathrm{t}\cdot\mathrm{h}$ & $\mathrm{p}\cdot\mathrm{t}$ & Share (\%) & $E_{\rm pos}$ & $E_{\rm twin}$ & $E_{\rm drift}$ \\
\midrule
0.05 & 0.65 & 0.02 & 0.25 & 6.3 & \phantom{0}167 & \phantom{0}47 & \phantom{0}163 \\
0.25 & 0.68 & 0.05 & 0.23 & 5.3 & \phantom{0}122 & \phantom{0}37 & \phantom{0}118 \\
0.45 & 0.69 & 0.08 & 0.21 & 4.3 & \phantom{0}133 & \phantom{0}37 & \phantom{0}125 \\
0.65 & 0.71 & 0.11 & 0.19 & 3.7 & \phantom{0}170 & \phantom{0}49 & \phantom{0}165 \\
0.75 & 0.72 & 0.16 & 0.19 & 3.6 & \phantom{0}245 & \phantom{0}72 & \phantom{0}240 \\
0.85 & 0.74 & 0.25 & 0.17 & 3.3 & \phantom{0}470 & 145 & \phantom{0}467 \\
\rowcolor{oursbg}
\textbf{0.95} & \textbf{0.79} & \textbf{0.35} & 0.14 & 2.9 & \textbf{2326} & 542 & \textbf{2270} \\
\bottomrule
\end{fittab}
\end{table}

\begin{table}[t]
\centering
\footnotesize
\renewcommand{\arraystretch}{1.08}
\caption{\textbf{Velocity probe on FLUX.2-klein.} Medians across 302 first-round base-model failures at each noise level $\sigma$, with columns defined in \cref{tab:probe_qwen}.}
\label{tab:probe_flux}
\begin{fittab}{\linewidth}{@{}F{c} ccc c cc E{c}@{}}
\toprule
& \multicolumn{3}{c}{Directional cosine} & & \multicolumn{3}{c@{}}{Squared norm} \\
\cmidrule(lr){2-4}\cmidrule(l){6-8}
$\sigma$ & $\mathrm{p}\cdot\mathrm{d}$ & $\mathrm{t}\cdot\mathrm{h}$ & $\mathrm{p}\cdot\mathrm{t}$ & Share (\%) & $E_{\rm pos}$ & $E_{\rm twin}$ & $E_{\rm drift}$ \\
\midrule
0.05 & 0.20 & 0.01 & 0.43 & 18.4 & \phantom{00}82 & \phantom{0}71 & \phantom{00}82 \\
0.25 & 0.10 & 0.01 & 0.46 & 21.6 & \phantom{0}284 & 256 & \phantom{0}283 \\
0.45 & 0.08 & 0.01 & 0.47 & 22.2 & \phantom{0}338 & 313 & \phantom{0}333 \\
0.65 & 0.09 & 0.01 & 0.46 & 21.6 & \phantom{0}351 & 314 & \phantom{0}346 \\
0.75 & 0.12 & 0.03 & 0.45 & 20.1 & \phantom{0}350 & 304 & \phantom{0}349 \\
0.85 & 0.28 & 0.07 & 0.39 & 14.9 & \phantom{0}422 & 324 & \phantom{0}422 \\
\rowcolor{oursbg}
\textbf{0.95} & \textbf{0.76} & \textbf{0.46} & 0.16 & \phantom{0}3.1 & \textbf{1520} & 549 & \textbf{1503} \\
\bottomrule
\end{fittab}
\end{table}

\paragraph{Qwen-Image-2512.} \cref{tab:probe_qwen,fig:motivation_supp}a repeat the velocity probe of \cref{subsec:drift} on 400 first-round base-model failures of Qwen-Image-2512. Across all noise levels, the full-revision update $\Delta_{\rm pos}$ closely follows the rewrite drift $\Delta_{\rm drift}$ (median cosine 0.65 to 0.79), leaving only 2.9\% to 3.7\% of its energy along $\Delta_{\rm twin}$ at high noise ($\sigma\ge0.65$). Because most of $E_{\rm pos}$ comes from the scene rewrite ($E_{\rm drift}=118$ to $2270$), canceling that shared rewrite in $\Delta_{\rm twin}$ cuts the update's squared $L_2$ norm from $E_{\rm pos}$ down to $E_{\rm twin}$, making it three to four times smaller at every noise level (\eg, from 122 to 37 at $\sigma=0.25$, and from 2326 to 542 at $\sigma=0.95$). This smaller, drift-free update lets us scale $\Delta_{\rm twin}$ up safely in \cref{eq:target}. Meanwhile, agreement between the two local fixes $\Delta_{\rm twin}$ and $\Delta_{\rm hind}$ increases with noise from 0.02 at $\sigma=0.05$ to 0.35 at $\sigma=0.95$ (above 0.5 on 34\% of failures); absolute cosines are lower than on SD3.5-M (\cref{fig:motivation_supp}b) because Qwen-Image-2512's $1024^2$ latent space has four times as many dimensions.

\paragraph{FLUX.2-klein.} \cref{tab:probe_flux,fig:motivation_supp}c run the same probe on 302 first-round failures of FLUX.2-klein, where the effect concentrates much more sharply at high noise. For $\sigma\le0.85$, prompt differences alter the velocity field much less ($E_{\rm pos}\le422$), $\Delta_{\rm pos}$ has low cosine with $\Delta_{\rm drift}$ (0.08 to 0.28), and the two local fixes $\Delta_{\rm twin}$ and $\Delta_{\rm hind}$ barely agree (median cosine 0.01 to 0.07). Near pure noise ($\sigma=0.95$), where global layout is set, $E_{\rm pos}$ jumps nearly fourfold to 1520 and is almost entirely rewrite drift ($E_{\rm drift}=1503$; 0.76 cosine with $\Delta_{\rm pos}$, with only 3.1\% of $E_{\rm pos}$ along $\Delta_{\rm twin}$). Canceling the shared rewrite in $\Delta_{\rm twin}$ cuts this squared norm nearly threefold from $E_{\rm pos}=1520$ down to $E_{\rm twin}=549$, while agreement between $\Delta_{\rm twin}$ and $\Delta_{\rm hind}$ rises sharply to 0.46 (above 0.5 on 45\% of failures). Because local fixes on FLUX.2-klein only align near $\sigma=0.95$, weighting updates toward high noise with $w(\sigma)=2\sigma$ in \cref{eq:target} is essential.

\section{Evaluation Details}
\label{sec:supp_eval}

\begin{table*}[t]
\centering
\footnotesize
\renewcommand{\arraystretch}{1.10}
\caption{\textbf{Alignment and style takeover under habit-carrying critics.} Extending \cref{fig:motivation}b and \cref{fig:habitgrid}, we report three-seed mean $\pm$ std under current weights for Soft-TIFA GM and the probe rate of the injected style (Hab., \%), after four rounds on Qwen-Image-2512 (full 800-prompt protocol) and six rounds on SD3.5-M and FLUX.2-klein (held-out prompts). \textbf{Bold} marks the best trained mean per row.}
\label{tab:habit}
\newcolumntype{O}{>{\columncolor{oursbg}[\tabcolsep][\dimexpr\tabcolsep+0.5pt\relax]}c}
\newcolumntype{P}{>{\columncolor{oursbg}[\dimexpr\tabcolsep+0.5pt\relax][0pt]}c}
\begin{fittab}{\textwidth}{@{}F{l} l cc cc cc OP@{}}
\toprule
& & \multicolumn{2}{c}{Base model} & \multicolumn{2}{c}{Positive teacher} & \multicolumn{2}{c}{Cross-author pair}
& \multicolumn{2}{c@{}}{\textbf{\method (ours)}} \\
\cmidrule(lr){3-4}\cmidrule(lr){5-6}\cmidrule(lr){7-8}\cmidrule(l){9-10}
Backbone & Critic & GM $\uparrow$ & Hab. $\downarrow$ & GM $\uparrow$ & Hab. $\downarrow$ & GM $\uparrow$ & Hab. $\downarrow$ & \multicolumn{1}{c}{GM $\uparrow$} & \multicolumn{1}{E{c}@{}}{Hab. $\downarrow$} \\
\midrule
Qwen-Image-2512 & Sunset light & 31.8 & \phantom{0}1.4 & 31.2\sd{1.0} & 87.1\sd{3.6} & 37.6\sd{1.5} & 98.6\sd{0.2} & \bestpm{49.3}{1.6} & \phantom{0}\bestpm{1.2}{0.1} \\
 & Cold light & 31.8 & 13.4 & 31.1\sd{2.1} & 87.3\sd{2.7} & 36.9\sd{1.4} & 99.1\sd{0.3} & \bestpm{48.2}{1.7} & \bestpm{14.2}{2.0} \\
\midrule
SD3.5-M & Sunset light & 19.6 & \phantom{0}2.2 & 19.4\sd{0.7} & 82.1\sd{2.6} & 16.0\sd{2.1} & 99.6\sd{0.1} & \bestpm{30.2}{0.5} & \phantom{0}\bestpm{2.4}{0.2} \\
 & Cold light & 19.6 & \phantom{0}7.2 & 20.6\sd{1.4} & 52.3\sd{3.4} & \bestpm{24.0}{1.5} & 57.3\sd{5.5} & 22.7\sd{0.3} & \phantom{0}\bestpm{7.5}{0.1} \\
 & Vintage film & 19.6 & \phantom{0}2.6 & 22.0\sd{0.9} & 84.4\sd{2.4} & 11.4\sd{1.8} & 99.9\sd{0.0} & \bestpm{31.2}{2.4} & \phantom{0}\bestpm{2.9}{1.0} \\
\midrule
FLUX.2-klein & Sunset light & 23.5 & \phantom{0}2.7 & 22.8\sd{1.3} & 85.0\sd{1.9} & \bestpm{26.7}{1.9} & 89.9\sd{6.3} & 25.4\sd{1.1} & \phantom{0}\bestpm{2.1}{0.3} \\
 & Cold light & 23.5 & 11.3 & 21.9\sd{1.0} & 89.2\sd{2.8} & \bestpm{26.9}{1.3} & 92.2\sd{2.3} & 24.0\sd{2.1} & \phantom{0}\bestpm{9.8}{0.8} \\
 & Vintage film & 23.5 & \phantom{0}3.8 & 25.6\sd{0.2} & 93.4\sd{5.6} & \bestpm{28.5}{1.6} & 94.2\sd{6.6} & 24.4\sd{1.3} & \phantom{0}\bestpm{2.7}{0.2} \\
\bottomrule
\end{fittab}
\end{table*}

\begin{table*}[t]
\centering
\footnotesize
\renewcommand{\arraystretch}{1.00}
\caption{\textbf{Full habit-probe rates across backbones and critics.} On the checkpoints from \cref{tab:heldout,tab:habit}, we report mean $\pm$ std probe rates (\%) for plain backgrounds (Plain), sunset light (Sun.), cold light (Cold), and vintage film (Vint.; ``---'' when not evaluated). Test-time rev.\ (1 pass) applies one critic-revision pass to the base model without training.}
\label{tab:fingerprint}
\begin{fittab}{\textwidth}{@{}F{l} l l c c c E{c}@{}}
\toprule
Backbone & Critic & Method & Plain (\%) & Sun. (\%) & Cold (\%) & Vint. (\%) \\
\midrule
Qwen-Image-2512 & Default & \cellcolor{basebg}Base model & \cellcolor{basebg}21.0\sdz{0.0} & \cellcolor{basebg}\phantom{0}1.4\sdz{0.0} & \cellcolor{basebg}13.4\sdz{0.0} & \cellcolor{basebg}--- \\
 & & Positive teacher & 22.7\sd{1.1} & \phantom{0}1.0\sd{0.0} & 15.7\sd{0.6} & --- \\
 & & Cross-author pair & 15.4\sd{5.0} & \phantom{0}0.8\sd{0.1} & 21.2\sd{4.3} & 10.1\sdz{0.0} \\
 & & Test-time rev.\ (1 pass) & 20.3\sdz{0.0} & \phantom{0}1.1\sdz{0.0} & 14.0\sdz{0.0} & --- \\
 & & \cellcolor{oursbg}\textbf{\method (ours)} & \cellcolor{oursbg}29.1\sd{1.9} & \cellcolor{oursbg}\phantom{0}0.8\sd{0.1} & \cellcolor{oursbg}13.1\sd{1.2} & \cellcolor{oursbg}--- \\
\cmidrule(l){2-7}
 & Sunset & Positive teacher & \phantom{0}4.6\sd{1.3} & 87.1\sd{3.6} & \phantom{0}0.3\sd{0.1} & --- \\
 & & Cross-author pair & \phantom{0}0.7\sd{0.3} & 98.6\sd{0.2} & \phantom{0}0.0\sd{0.0} & --- \\
 & & Test-time rev.\ (1 pass) & \phantom{0}9.7\sdz{0.0} & 64.9\sdz{0.0} & \phantom{0}4.2\sdz{0.0} & --- \\
 & & \cellcolor{oursbg}\textbf{\method (ours)} & \cellcolor{oursbg}27.1\sd{3.9} & \cellcolor{oursbg}\phantom{0}1.2\sd{0.1} & \cellcolor{oursbg}12.0\sd{0.5} & \cellcolor{oursbg}--- \\
\cmidrule(l){2-7}
 & Cold & Positive teacher & 14.4\sd{2.6} & \phantom{0}0.1\sd{0.0} & 87.3\sd{2.7} & 22.3\sdz{0.0} \\
 & & Cross-author pair & \phantom{0}5.9\sd{1.7} & \phantom{0}0.0\sd{0.0} & 99.1\sd{0.3} & 32.1\sd{8.0} \\
 & & \cellcolor{oursbg}\textbf{\method (ours)} & \cellcolor{oursbg}23.2\sd{0.5} & \cellcolor{oursbg}\phantom{0}1.0\sd{0.4} & \cellcolor{oursbg}14.2\sd{2.0} & \cellcolor{oursbg}\phantom{0}2.0\sdz{0.0} \\
\midrule
SD3.5-M & Default & \cellcolor{basebg}Base model & \cellcolor{basebg}62.2\sdz{0.0} & \cellcolor{basebg}\phantom{0}2.2\sdz{0.0} & \cellcolor{basebg}\phantom{0}7.2\sdz{0.0} & \cellcolor{basebg}\phantom{0}2.6\sdz{0.0} \\
 & & Positive teacher & 48.2\sd{6.3} & \phantom{0}3.2\sd{0.5} & \phantom{0}8.5\sd{1.0} & --- \\
 & & Cross-author pair & 49.8\sd{6.3} & \phantom{0}3.7\sd{1.0} & \phantom{0}9.7\sd{2.2} & \phantom{0}2.5\sdz{0.0} \\
 & & \cellcolor{oursbg}\textbf{\method (ours)} & \cellcolor{oursbg}76.2\sd{3.6} & \cellcolor{oursbg}\phantom{0}1.5\sd{0.2} & \cellcolor{oursbg}\phantom{0}5.1\sd{1.2} & \cellcolor{oursbg}--- \\
 & & \cellcolor{oursbg}\textbf{\method (round 9)} & \cellcolor{oursbg}82.9\sd{2.4} & \cellcolor{oursbg}\phantom{0}1.5\sd{1.0} & \cellcolor{oursbg}\phantom{0}5.3\sd{2.1} & \cellcolor{oursbg}--- \\
\cmidrule(l){2-7}
 & Sunset & Positive teacher & 11.1\sd{1.7} & 82.1\sd{2.6} & \phantom{0}1.2\sd{0.2} & \phantom{0}7.7\sd{1.7} \\
 & & Cross-author pair & \phantom{0}5.4\sd{1.3} & 99.6\sd{0.1} & \phantom{0}0.1\sd{0.0} & 29.1\sd{0.3} \\
 & & \cellcolor{oursbg}\textbf{\method (ours)} & \cellcolor{oursbg}71.9\sd{7.3} & \cellcolor{oursbg}\phantom{0}2.4\sd{0.2} & \cellcolor{oursbg}\phantom{0}6.2\sd{1.6} & \cellcolor{oursbg}\phantom{0}1.6\sd{0.9} \\
\cmidrule(l){2-7}
 & Cold & Positive teacher & 30.1\sd{3.5} & \phantom{0}1.5\sd{0.2} & 52.3\sd{3.4} & 12.7\sd{1.8} \\
 & & Cross-author pair & 27.9\sd{4.7} & \phantom{0}0.9\sd{0.2} & 57.3\sd{5.5} & 13.6\sd{2.7} \\
 & & \cellcolor{oursbg}\textbf{\method (ours)} & \cellcolor{oursbg}63.2\sd{1.3} & \cellcolor{oursbg}\phantom{0}2.0\sd{0.1} & \cellcolor{oursbg}\phantom{0}7.5\sd{0.1} & \cellcolor{oursbg}\phantom{0}2.4\sd{0.5} \\
\cmidrule(l){2-7}
 & Vintage & Positive teacher & 23.1\sd{7.6} & \phantom{0}4.0\sd{1.7} & 21.7\sd{3.1} & 84.4\sd{2.4} \\
 & & Cross-author pair & 16.5\sd{13.3} & 17.6\sd{9.0} & 10.8\sd{3.6} & 99.9\sd{0.0} \\
 & & \cellcolor{oursbg}\textbf{\method (ours)} & \cellcolor{oursbg}66.5\sd{0.7} & \cellcolor{oursbg}\phantom{0}2.0\sd{0.4} & \cellcolor{oursbg}\phantom{0}7.4\sd{0.6} & \cellcolor{oursbg}\phantom{0}2.9\sd{1.0} \\
\midrule
FLUX.2-klein & Default & \cellcolor{basebg}Base model & \cellcolor{basebg}51.4\sdz{0.0} & \cellcolor{basebg}\phantom{0}2.7\sdz{0.0} & \cellcolor{basebg}11.3\sdz{0.0} & \cellcolor{basebg}\phantom{0}3.8\sdz{0.0} \\
 & & Positive teacher & 40.4\sd{1.5} & \phantom{0}1.8\sd{0.5} & 13.3\sd{1.2} & --- \\
 & & Cross-author pair & 46.6\sd{6.1} & \phantom{0}1.7\sd{1.1} & \phantom{0}4.6\sd{2.1} & --- \\
 & & \cellcolor{oursbg}\textbf{\method (ours)} & \cellcolor{oursbg}57.0\sd{11.7} & \cellcolor{oursbg}\phantom{0}2.2\sd{0.4} & \cellcolor{oursbg}\phantom{0}9.6\sd{1.6} & \cellcolor{oursbg}--- \\
\cmidrule(l){2-7}
 & Sunset & Positive teacher & 24.5\sd{6.3} & 85.0\sd{1.9} & \phantom{0}0.3\sd{0.0} & \phantom{0}9.5\sd{1.8} \\
 & & Cross-author pair & 23.1\sd{10.2} & 89.9\sd{6.3} & \phantom{0}0.2\sd{0.0} & \phantom{0}5.2\sd{0.2} \\
 & & \cellcolor{oursbg}\textbf{\method (ours)} & \cellcolor{oursbg}55.7\sd{9.5} & \cellcolor{oursbg}\phantom{0}2.1\sd{0.3} & \cellcolor{oursbg}10.5\sd{1.5} & \cellcolor{oursbg}\phantom{0}3.6\sd{0.9} \\
\cmidrule(l){2-7}
 & Cold & Positive teacher & 50.1\sd{16.2} & \phantom{0}0.2\sd{0.0} & 89.2\sd{2.8} & 24.8\sd{5.9} \\
 & & Cross-author pair & 48.7\sd{5.1} & \phantom{0}0.2\sd{0.0} & 92.2\sd{2.3} & 22.8\sd{3.1} \\
 & & \cellcolor{oursbg}\textbf{\method (ours)} & \cellcolor{oursbg}54.0\sd{3.1} & \cellcolor{oursbg}\phantom{0}2.1\sd{0.2} & \cellcolor{oursbg}\phantom{0}9.8\sd{0.8} & \cellcolor{oursbg}\phantom{0}3.3\sd{0.5} \\
\cmidrule(l){2-7}
 & Vintage & Positive teacher & 42.6\sd{6.6} & 19.2\sd{3.6} & \phantom{0}7.2\sd{2.1} & 93.4\sd{5.6} \\
 & & Cross-author pair & 36.5\sd{2.3} & 17.3\sd{1.1} & \phantom{0}6.2\sd{1.1} & 94.2\sd{6.6} \\
 & & \cellcolor{oursbg}\textbf{\method (ours)} & \cellcolor{oursbg}57.4\sd{6.5} & \cellcolor{oursbg}\phantom{0}2.1\sd{0.2} & \cellcolor{oursbg}10.5\sd{1.1} & \cellcolor{oursbg}\phantom{0}2.7\sd{0.2} \\
\bottomrule
\end{fittab}
\end{table*}

\begin{figure}[b]
  \centering
  \includegraphics[width=\linewidth]{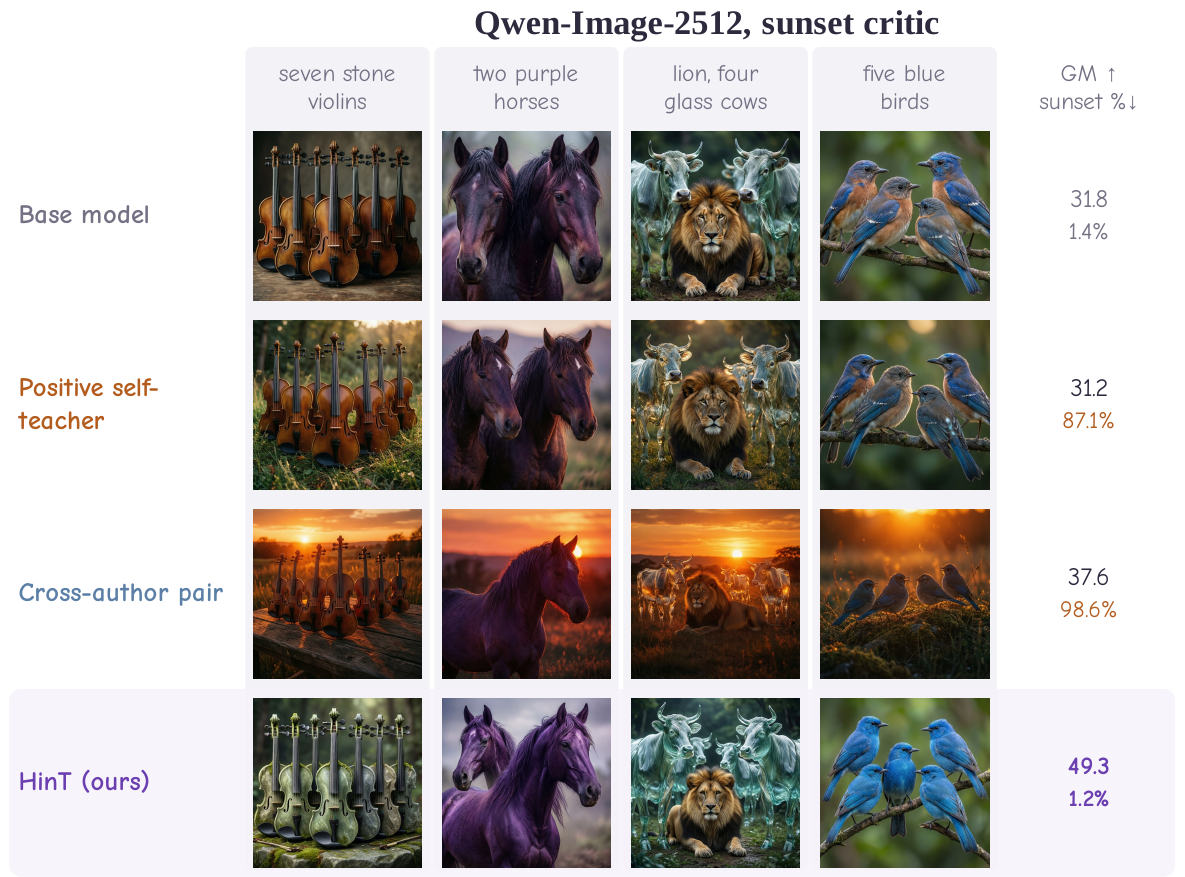}
  \caption{\textbf{Habit transfer on Qwen-Image-2512 under the sunset critic.} Complementing \cref{fig:habitgrid}, each column shares a test prompt and seed after four training rounds; the rightmost column lists three-seed mean GM and sunset rate from \cref{tab:habit}. Both baselines cast every scene in warm sunset light, whereas \method keeps neutral lighting and fixes the count and material errors.}
  \label{fig:habitgrid_qwen}
\end{figure}

\paragraph{Benchmark judges.} On GenEval2, we report Soft-TIFA geometric mean (GM) and arithmetic mean (AM) from the official Qwen3-VL-8B judge~\cite{kamath2025geneval2}, alongside J2 from an independent InternVL3.5-38B judge~\cite{wang2025internvl35} that answers the exact same atomic questions and aggregates them in the same way. Unless noted otherwise, single-column judge and quality metrics (J2 in \cref{tab:heldout,tab:heldout_full,tab:rounds} and \cref{tab:quality}) evaluate current weights on Qwen-Image-2512 (or whenever student EMA is unused) and EMA weights on SD3.5-M and FLUX.2-klein. GenEval uses its official Mask2Former detector, and PickScore in \cref{tab:qwen} is averaged over prompt--image pairs. On OneIG-Bench, \cref{tab:ood,tab:ood_full} compute mean dependency-gated question accuracy separately with Qwen3-VL-8B and InternVL3.5-38B (J2) for cross-benchmark consistency, while \cref{tab:oneig_official} runs OneIG-Bench's official Qwen2.5-VL-7B scorer and confirms the same ranking.

\paragraph{Precision and evaluation variance.} During training, checkpoints are scored online with one image per prompt (seed 900); after training, standalone evaluation loads bf16 LoRA adapters in larger batches and averages four images per prompt (seeds 900 to 903). Even on the shared seed 900, bf16 rounding and different batching perturb the sampled images, yet the two single-image scores agree on average across identical checkpoints ($-0.15$\sd{1.81} GM over 82 SD3.5-M runs and $+0.35$\sd{3.35} GM over 40 FLUX.2-klein runs), ruling out bf16 bias. Treating those two perturbed single-image runs as independent draws ($\mathrm{Var}=2\sigma_1^2$) gives a single-image noise of $\sigma_1\approx1.3$ GM on SD3.5-M and $2.4$ GM on FLUX.2-klein, so averaging four seeds per prompt reduces single-checkpoint standard error to 0.6 and 1.2 GM.

\paragraph{Habit probe and style-transfer results.} To measure unrequested visual styles independently of the user prompt, InternVL3.5-38B answers four yes/no questions on every evaluation image (and on the 400 round-start training rollouts per round in \cref{fig:motivation}c): (1) ``Is the scene lit by warm, golden sunset light?''; (2) ``Is the scene lit by cold, blue, overcast light?''; (3) ``Is the background a plain, uniform color with no scenery?''; and (4) ``Does the image look like a vintage film photograph, with faded colors and visible film grain?''. \cref{tab:habit,tab:fingerprint} report the mean ``yes'' probability in percent. Under the habit-carrying critics (\cref{tab:habit,fig:habitgrid_qwen}), both baselines adopt the injected lighting or film look on 52.3\% to 99.9\% of test images across all eight backbone--critic settings, whereas \method stays near the base model's natural rate throughout. Even under the default critic with no style instruction (\cref{tab:fingerprint}), critic-added scenery lowers plain-background rates for both baselines on SD3.5-M (from 62.2\% base to 48.2\% and 49.8\%) and FLUX.2-klein (from 51.4\% to 40.4\% and 46.6\%), and for the cross-author pair on Qwen-Image-2512 (from 21.0\% to 15.4\%), whereas \method preserves clean backgrounds across all three backbones (29.1\%, 76.2\%, and 57.0\%; \cref{subsec:fingerprints}).

\paragraph{Validating the habit-transfer setup.} Two control checks rule out potential confounds on both the critic side and the rater side:
\begin{itemize}[leftmargin=1.2em, itemsep=1pt, topsep=2pt]
  \item \textbf{Style instructions do not alter error diagnosis.} Because round-1 rollouts come from the frozen base model with fixed seeds, different critics can be compared on identical images. Across the 800 Qwen-Image-2512 rollouts, the sunset and cold-light critics agree with the default critic on 98.9\% to 99.1\% of pass/fail decisions, 99.2\% of the 6,012 atomic questions, and 92.4\% to 92.6\% of failed spans and fixes, while adding the requested lighting to 100\% of rejected templates (vs.\ 0\% under the default critic). Across the 400 SD3.5-M rollouts, the sunset and vintage-film critics likewise agree with the default critic on 99.0\% to 99.3\% of pass/fail decisions and 98.2\% to 98.5\% of atomic questions. Style instructions thus change only the surrounding template $T$, not which errors the critic diagnoses.
  \item \textbf{Habit rankings do not depend on the rater.} Re-scoring 12,000 test images across ten evaluation sets with Qwen3-VL-8B yields close agreement with InternVL3.5-38B on both sunset light ($r=0.98$ per-image correlation, 98.4\% binary agreement, and set means differing by at most 2.4 points) and plain backgrounds ($r=0.92$ correlation and 93.7\% binary agreement), leaving the method ordering unchanged on every set.
\end{itemize}

\makeatletter\def\ftype@figure{2}\makeatother
\section{Additional Results}
\label{sec:supp_results}

\subsection{Extended GenEval2 and Multi-Round Results}

\begin{table*}[t]
\centering
\footnotesize
\renewcommand{\arraystretch}{1.08}
\caption{\textbf{Held-out GenEval2 results with seed standard deviations.} Expanding \cref{tab:heldout}, we report mean $\pm$ std across seeds for Soft-TIFA GM (EMA and Cur.) and J2 (\cref{sec:supp_eval}); single-run cells omit $\pm$. \textbf{Bold} marks the best mean per column.}
\label{tab:heldout_full}
\begin{fittab}{\textwidth}{@{}F{l} ccc ccc ccE{c}@{}}
\toprule
& \multicolumn{3}{c}{Qwen-Image-2512} & \multicolumn{3}{c}{SD3.5-M} & \multicolumn{3}{c@{}}{FLUX.2-klein} \\
\cmidrule(lr){2-4}\cmidrule(lr){5-7}\cmidrule(l){8-10}
Method & EMA & Cur. & J2 & EMA & Cur. & J2 & EMA & Cur. & J2 \\
\midrule
\rowcolor{basebg}
Base model & \multicolumn{2}{c}{\cellcolor{basebg}33.9} & 50.4\sdz{0.0} & \multicolumn{2}{c}{\cellcolor{basebg}19.6} & 35.4\sdz{0.0} & \multicolumn{2}{c}{\cellcolor{basebg}23.5} & 40.9\sdz{0.0} \\
Positive teacher & 33.5\sd{0.2} & 34.8\sd{0.6} & 51.5\sd{0.8} & --- & 21.4\sd{0.4} & 38.8\sd{0.9} & --- & 20.0\sd{0.6} & 36.5\sd{1.3} \\
SFT on revised prompt & --- & 38.6\sd{1.1} & 54.1\sd{0.3} & --- & 21.8\sdz{0.0} & 39.2\sdz{0.0} & --- & \phantom{0}8.6\sdz{0.0} & 29.5\sdz{0.0} \\
DiffusionNFT & 33.3\sd{1.3} & 33.8\sd{1.9} & 50.4\sd{1.8} & 19.8\sd{1.0} & 20.0\sd{1.7} & 36.0\sd{1.3} & 23.0\sd{0.4} & 22.5\sd{0.7} & 40.5\sd{0.5} \\
Cross-author pair & 40.9\sd{1.0} & 41.9\sd{1.3} & 56.5\sd{0.8} & 29.1\sd{1.7} & \bestpm{28.5}{2.3} & 45.8\sd{2.4} & \bestpm{27.3}{3.7} & 22.4\sd{6.3} & \bestpm{45.8}{4.0} \\
User hindsight & 43.2\sd{1.5} & 42.0\sd{2.0} & 57.2\sd{2.1} & 24.8\sd{3.3} & 24.7\sd{4.7} & 41.1\sd{3.8} & 10.4\sd{9.0} & \phantom{0}5.4\sd{4.0} & 25.8\sd{15.9} \\
\rowcolor{oursbg}
\textbf{\method (ours)} & \bestpm{45.3}{1.1} & \bestpm{44.7}{2.3} & \bestpm{59.2}{2.2} & \bestpm{30.7}{1.1} & 27.5\sd{1.7} & \bestpm{47.9}{1.2} & 25.4\sd{1.7} & \bestpm{26.3}{0.7} & 43.9\sd{2.0} \\
\bottomrule
\end{fittab}
\end{table*}

\begin{table*}[t]
\centering
\footnotesize
\renewcommand{\arraystretch}{1.08}
\caption{\textbf{Multi-round results at 6, 9, and 12 training rounds.} Extending \cref{fig:rounds}, we report mean $\pm$ std across seeds on the full 800-prompt protocol for Qwen-Image-2512 and on held-out prompts for SD3.5-M and FLUX.2-klein (``---'' indicates runs that collapsed by round 6). \textbf{Bold} marks the best mean at each round.}
\label{tab:rounds}
\newcolumntype{O}{>{\columncolor{oursbg}[\tabcolsep][\dimexpr\tabcolsep+0.5pt\relax]}c}
\newcolumntype{Q}{>{\columncolor{oursbg}[\dimexpr\tabcolsep+0.5pt\relax][\dimexpr\tabcolsep+0.5pt\relax]}c}
\newcolumntype{P}{>{\columncolor{oursbg}[\dimexpr\tabcolsep+0.5pt\relax][0pt]}c}
\begin{fittab}{\textwidth}{@{}F{l} c ccc ccc OQP@{}}
\toprule
& & \multicolumn{3}{c}{Cross-author pair} & \multicolumn{3}{c}{User hindsight} & \multicolumn{3}{c@{}}{\textbf{\method (ours)}} \\
\cmidrule(lr){3-5}\cmidrule(lr){6-8}\cmidrule(l){9-11}
Backbone & Rounds & EMA & Cur. & J2 & EMA & Cur. & J2 & \multicolumn{1}{c}{EMA} & \multicolumn{1}{c}{Cur.} & \multicolumn{1}{E{c}@{}}{J2} \\
\midrule
Qwen-Image-2512 & \phantom{0}6 & 46.8\sd{0.7} & 48.4\sd{0.4} & 61.8\sd{0.4} & 47.7\sd{0.9} & 48.2\sd{2.0} & 61.3\sd{0.7} & \bestpm{53.0}{0.4} & \bestpm{54.1}{2.5} & \bestpm{65.9}{1.4} \\
 & \phantom{0}9 & 48.6\sd{2.9} & 47.7\sd{4.3} & 61.1\sd{3.7} & 47.2\sd{4.6} & 47.0\sd{4.4} & 61.0\sd{4.1} & \bestpm{55.3}{2.7} & \bestpm{53.9}{0.8} & \bestpm{65.1}{1.8} \\
 & 12 & 44.3\sd{4.7} & 42.9\sd{5.0} & 57.1\sd{5.4} & 46.6\sd{5.4} & 45.4\sd{6.0} & 60.0\sd{5.2} & \bestpm{56.4}{1.4} & \bestpm{54.7}{1.1} & \bestpm{67.1}{1.1} \\
\midrule
SD3.5-M & \phantom{0}6 & 29.1\sd{1.7} & \bestpm{28.5}{2.3} & 45.8\sd{2.4} & 24.8\sd{3.3} & 24.7\sd{4.7} & 41.1\sd{3.8} & \bestpm{30.7}{1.1} & 27.5\sd{1.7} & \bestpm{47.9}{1.2} \\
 & \phantom{0}9 & 30.0\sd{1.5} & 29.9\sd{1.7} & 47.5\sd{1.4} & 22.9\sd{4.0} & 21.0\sd{1.7} & 39.7\sd{5.1} & \bestpm{31.4}{1.0} & \bestpm{30.9}{1.4} & \bestpm{49.5}{1.7} \\
 & 12 & 29.3\sd{2.2} & 28.6\sd{1.6} & \bestpm{47.9}{1.8} & 15.9\sd{0.6} & 12.9\sd{3.1} & 33.3\sd{0.5} & \bestpm{29.7}{0.4} & \bestpm{29.5}{0.7} & 47.4\sd{1.8} \\
\midrule
FLUX.2-klein & \phantom{0}6 & \bestpm{27.3}{3.7} & 22.4\sd{6.3} & \bestpm{45.8}{4.0} & --- & --- & --- & 25.4\sd{1.7} & \bestpm{26.3}{0.7} & 43.9\sd{2.0} \\
 & \phantom{0}9 & 30.5\sd{1.8} & 28.4\sd{3.6} & 49.8\sd{2.0} & --- & --- & --- & \bestpm{30.6}{1.3} & \bestpm{30.6}{0.7} & \bestpm{50.5}{0.8} \\
 & 12 & 29.3\sd{1.0} & 27.2\sd{0.2} & 49.0\sd{1.6} & --- & --- & --- & \bestpm{31.4}{1.2} & \bestpm{31.3}{1.6} & \bestpm{51.5}{1.3} \\
\bottomrule
\end{fittab}
\end{table*}

\begin{table*}[t]
\centering
\footnotesize
\renewcommand{\arraystretch}{1.08}
\caption{\textbf{Extended GenEval2 comparison on Qwen-Image-2512.} All rows are evaluated on the 800 GenEval2 benchmark prompts (\cref{sec:supp_impl}). Trained models run for four rounds under two prompt setups: the top block trains on 800 disjoint grammar prompts (\cref{tab:qwen}), whereas the full 800-prompt protocol~\cite{wu2026unievo} uses the 800 benchmark prompts themselves as unlabeled training prompts (supervised only by VLM critiques on sampled rollouts), tested at 1 vs.\ 4 rollouts per prompt and with test-time critic revision (regenerating failed test images from $r$ for 1 or 2 passes on the base model or round-4 checkpoints). GM, AM, J2, and PickScore (Pick) use current weights, EMA reports EMA-weight GM, cells without $\pm$ are single runs, and \textbf{bold} marks the best mean per block.}
\label{tab:qwen_full}
\begin{fittab}{\textwidth}{@{}F{l} c c c c E{c}@{}}
\toprule
Method & GM $\uparrow$ & AM $\uparrow$ & J2 $\uparrow$ & Pick $\uparrow$ & EMA $\uparrow$ \\
\midrule
\multicolumn{6}{@{}l}{\textit{Disjoint-prompt protocol (\cref{tab:qwen}; trained on 800 separate grammar prompts)}} \\
\rowcolor{basebg}
Base model & 31.8\sdz{0.0} & 78.3\sdz{0.0} & 48.8\sdz{0.0} & 23.86\sdz{0.00} & --- \\
DiffusionNFT & 27.4\sd{1.6} & 76.4\sd{0.1} & 44.6\sd{1.9} & 23.78\sd{0.02} & 30.4\sd{1.1} \\
Positive teacher & 33.9\sd{1.2} & 78.6\sd{0.6} & 51.3\sd{1.0} & 23.83\sd{0.02} & 32.6\sd{0.8} \\
Cross-author pair & 44.4\sd{2.1} & 82.2\sd{0.6} & 58.7\sd{1.0} & 23.79\sd{0.03} & 42.5\sd{1.4} \\
User hindsight & 43.5\sd{2.6} & 82.8\sd{1.0} & 57.6\sd{2.2} & 23.92\sd{0.03} & 43.0\sd{3.5} \\
\rowcolor{oursbg}
\textbf{\method (ours)} & \bestpm{49.0}{1.8} & \bestpm{84.2}{0.6} & \bestpm{62.8}{1.7} & \bestpm{23.99}{0.04} & \bestpm{47.2}{2.6} \\
\midrule
\multicolumn{6}{@{}l}{\textit{Full 800-prompt protocol~\cite{wu2026unievo} (trained on the 800 GenEval2 prompts): single rollout per prompt}} \\
\rowcolor{basebg}
Base model & 31.8\sdz{0.0} & 78.3\sdz{0.0} & 48.8\sdz{0.0} & 23.86\sdz{0.00} & --- \\
DiffusionNFT & 32.7\sd{4.5} & 78.6\sd{1.4} & 49.4\sd{3.6} & 23.81\sd{0.01} & --- \\
Positive teacher & 35.0\sd{1.5} & 79.0\sd{0.7} & 52.1\sd{1.3} & 23.80\sd{0.01} & 32.7\sd{0.3} \\
Cross-author pair & 44.5\sd{0.6} & 82.3\sd{0.1} & 59.0\sd{0.0} & 23.73\sd{0.03} & 43.1\sd{1.1} \\
User hindsight & 46.7\sd{0.4} & 83.9\sd{0.1} & 60.3\sd{0.3} & 23.95\sd{0.06} & 44.2\sd{1.4} \\
\rowcolor{oursbg}
\textbf{\method (ours)} & \bestpm{48.8}{4.4} & \bestpm{84.5}{1.5} & \bestpm{62.3}{3.3} & \bestpm{23.97}{0.04} & \bestpm{46.8}{4.5} \\
\midrule
\multicolumn{6}{@{}l}{\textit{Full 800-prompt protocol: four rollouts per prompt}} \\
DiffusionNFT & 37.8\sd{1.3} & 80.1\sd{0.2} & 53.5\sd{1.1} & 23.82\sd{0.04} & --- \\
\rowcolor{oursbg}
\textbf{\method (ours)} & \best{48.8}\sdz{0.0} & \best{84.8}\sdz{0.0} & \best{62.6}\sdz{0.0} & \best{24.01}\sdz{0.00} & --- \\
\midrule
\multicolumn{6}{@{}l}{\textit{Full 800-prompt protocol: test-time critic revision on failed outputs (seed 1)}} \\
Base model (1 pass) & 38.9\sdz{0.0} & 80.0\sdz{0.0} & 55.3\sdz{0.0} & 23.81\sdz{0.00} & --- \\
Base model (2 passes) & 41.5\sdz{0.0} & 80.7\sdz{0.0} & 56.6\sdz{0.0} & 23.79\sdz{0.00} & --- \\
Cross-author pair (1 pass) & 48.5\sdz{0.0} & 83.7\sdz{0.0} & 62.7\sdz{0.0} & 23.66\sdz{0.00} & --- \\
\rowcolor{oursbg}
\textbf{\method (1 pass)} & \best{59.4}\sdz{0.0} & \best{87.4}\sdz{0.0} & \best{69.6}\sdz{0.0} & \best{23.96}\sdz{0.00} & --- \\
\bottomrule
\end{fittab}
\end{table*}

\begin{table*}[t]
\centering
\footnotesize
\renewcommand{\arraystretch}{1.08}
\caption{\textbf{Per-skill GenEval2 accuracy on Qwen-Image-2512.} Under the full 800-prompt protocol of \cref{tab:qwen_full}, we report three-seed mean $\pm$ std atom accuracy (\%) across skills (Attr.: attribute; Pos.: position), with question counts in the top row. \textbf{Bold} marks the best trained mean per column.}
\label{tab:skills_full}
\begin{fittab}{\textwidth}{@{}F{l} ccccc ccccE{c}@{}}
\toprule
& \multicolumn{5}{c}{Official judge (Soft-TIFA)} & \multicolumn{5}{c@{}}{Independent judge (J2)} \\
\cmidrule(lr){2-6}\cmidrule(l){7-11}
Method & Attr. & Count & Object & Pos. & Verb & Attr. & Count & Object & Pos. & Verb \\
\midrule
Questions & 1214 & 2025 & 2025 & 662 & 86 & 1214 & 2025 & 2025 & 662 & 86 \\
\midrule
\rowcolor{basebg}
Base model & 74.6\sdz{0.0} & 66.7\sdz{0.0} & 98.8\sdz{0.0} & 61.4\sdz{0.0} & 33.9\sdz{0.0} & 76.0\sdz{0.0} & 65.1\sdz{0.0} & 99.1\sdz{0.0} & 65.4\sdz{0.0} & 40.5\sdz{0.0} \\
Positive teacher & 74.6\sd{0.1} & 69.3\sd{1.4} & 98.2\sd{0.2} & 60.4\sd{0.8} & 32.1\sd{0.9} & 76.1\sd{0.3} & 67.1\sd{1.2} & 98.8\sd{0.1} & 64.7\sd{0.5} & 39.2\sd{0.4} \\
Cross-author pair & 81.1\sd{0.5} & 74.5\sd{0.2} & 98.2\sd{0.1} & \bestpm{61.9}{0.3} & 35.2\sd{1.7} & 81.7\sd{0.5} & 70.7\sd{0.2} & 98.9\sd{0.2} & 65.9\sd{0.6} & 43.6\sd{0.8} \\
User hindsight & 86.2\sd{0.8} & \bestpm{74.7}{0.5} & \bestpm{98.6}{0.1} & 61.6\sd{0.8} & 46.1\sd{3.2} & 86.5\sd{0.6} & \bestpm{72.1}{0.5} & \bestpm{99.1}{0.1} & 66.5\sd{1.6} & \bestpm{54.1}{3.8} \\
\rowcolor{oursbg}
\textbf{\method (ours)} & \bestpm{89.8}{1.9} & 74.6\sd{2.9} & 98.3\sd{0.2} & 61.6\sd{0.8} & \bestpm{46.6}{3.4} & \bestpm{89.1}{1.5} & 72.0\sd{2.4} & 99.0\sd{0.2} & \bestpm{66.6}{0.6} & 51.6\sd{3.3} \\
\bottomrule
\end{fittab}
\end{table*}

\paragraph{Held-out and 800-prompt GenEval2.} \cref{tab:heldout_full} adds seed standard deviations to the held-out comparison in \cref{tab:heldout}, while \cref{tab:qwen_full} compares Qwen-Image-2512 on all 800 GenEval2 prompts after training either on 800 disjoint grammar prompts (\cref{tab:qwen}) or directly on the 800 benchmark prompts under the full 800-prompt protocol~\cite{wu2026unievo} (\cref{sec:supp_impl}). Under the full 800-prompt protocol at round four, LoRA EMA preserves the current-weight method ranking; one slower \method seed (44.6 GM) widens the round-four spread ($48.8$\sd{4.4} current, $46.8$\sd{4.5} EMA), but catches up by round six in \cref{tab:rounds} ($54.1$\sd{2.5} current, $53.0$\sd{0.4} EMA). Raising the rollout budget from one to four images per prompt helps DiffusionNFT through within-prompt group normalization (32.7 to 37.8 GM), yet single-rollout \method already reaches 48.8 GM and matches four-rollout \method (48.8 GM). At inference time, applying one pass of critic revision on failed outputs lifts the seed-1 \method checkpoint from 53.4 to 59.4 GM in \cref{tab:qwen_full}, outperforming one or two revision passes on the base model (38.9 and 41.5) and one pass on the seed-1 cross-author pair (44.2 to 48.5).

\paragraph{Multi-round stability and user-worded hindsight.} \cref{tab:rounds} reports the seed means and standard deviations behind \cref{fig:rounds} across rounds 6, 9, and 12. Consistent with the held-out trajectories in \cref{subsec:main}, zero-shot transfer on FLUX.2-klein also turns in \method's favor with longer training: although the cross-author pair starts higher on GenEval at round six (79.6 vs.\ 77.5 in \cref{tab:ood_full}), by round twelve it falls below the 76.4 base model to 72.6, whereas \method rises to 78.8 on GenEval and leads on OneIG-Bench (80.6 vs.\ 78.9). User-worded hindsight contrast ($\vel(c)-\vel(h)$) never uses $r$, avoiding style transfer by construction, and performs well on Qwen-Image-2512 (43.2 held-out and 46.7 800-prompt GM), whose Qwen2.5-VL text encoder responds strongly to short phrase edits. Without the critic's scene context $T$, however, contrasting inside the bare failed prompt $c$ is unstable on the other two backbones: on SD3.5-M it decays from $24.8$\sd{3.3} to $15.9$\sd{0.6} EMA GM between rounds 6 and 12 (\cref{tab:rounds}), and on FLUX.2-klein, where $\Delta_{\rm hind}$ stays nearly orthogonal to $\Delta_{\rm twin}$ below $\sigma=0.95$ (\cref{tab:probe_flux}), it collapses to $10.4$\sd{9.0} EMA GM by round six (\cref{tab:heldout_full}).

\subsection{Untouched Benchmarks and Image Quality}

\begin{table*}[t]
\centering
\footnotesize
\renewcommand{\arraystretch}{1.04}
\caption{\textbf{Zero-shot transfer to GenEval and OneIG-Bench.} Extending \cref{tab:ood}, we report seed mean $\pm$ std for GenEval overall accuracy ($\times100$, GE) and OneIG-Bench's General-Object (Obj.) and Portrait (Por.) subsets. Each OneIG score is the mean dependency-gated question accuracy (\%) across prompts, evaluated separately by Qwen3-VL-8B and InternVL3.5-38B (J2), with OneIG's official Qwen2.5-VL-7B scorer in \cref{tab:oneig_official}; single-seed cells omit $\pm$. \textbf{Bold} marks the best mean per backbone.}
\label{tab:ood_full}
\begin{fittab}{\textwidth}{@{}F{l} c cc cE{c}@{}}
\toprule
& & \multicolumn{2}{c}{OneIG (Qwen3-VL-8B)} & \multicolumn{2}{c@{}}{OneIG (InternVL3.5-38B, J2)} \\
\cmidrule(lr){3-4}\cmidrule(l){5-6}
Method & GE $\uparrow$ & Obj. $\uparrow$ & Por. $\uparrow$ & Obj. $\uparrow$ & Por. $\uparrow$ \\
\midrule
\multicolumn{6}{@{}l}{\textit{Qwen-Image-2512}} \\
\rowcolor{basebg}
Base model & 73.9\sdz{0.0} & 82.9\sdz{0.0} & \best{86.6}\sdz{0.0} & 89.2\sdz{0.0} & 92.2\sdz{0.0} \\
Positive teacher & 74.3\sd{0.3} & 82.9\sd{0.1} & 86.4\sd{0.2} & 89.2\sd{0.2} & 92.4\sd{0.1} \\
Cross-author pair & 75.2\sd{1.0} & 83.4\sd{0.5} & 86.1\sd{0.2} & 89.6\sd{0.2} & 92.5\sd{0.1} \\
User hindsight & \bestpm{77.9}{0.7} & 83.0\sd{0.4} & 85.1\sd{0.2} & 90.0\sdz{0.0} & 92.5\sdz{0.0} \\
\rowcolor{oursbg}
\textbf{\method (ours)} & 77.3\sd{0.7} & \bestpm{84.1}{0.4} & 86.0\sd{0.2} & \bestpm{90.3}{0.2} & \bestpm{92.6}{0.1} \\
\midrule
\multicolumn{6}{@{}l}{\textit{SD3.5-M}} \\
\rowcolor{basebg}
Base model & 67.3\sdz{0.0} & 73.1\sdz{0.0} & 78.9\sdz{0.0} & 83.2\sdz{0.0} & 88.0\sdz{0.0} \\
Positive teacher & 68.9\sd{0.0} & 74.6\sd{0.6} & 78.6\sd{0.2} & 84.4\sd{0.3} & 88.9\sd{0.2} \\
Cross-author pair & 73.5\sd{0.9} & 75.6\sd{0.2} & 79.2\sd{0.2} & 85.9\sd{0.4} & \bestpm{89.3}{0.2} \\
User hindsight & 72.1\sd{2.1} & 72.6\sd{2.4} & 76.7\sd{1.4} & 82.7\sd{2.1} & 87.2\sd{1.0} \\
\rowcolor{oursbg}
\textbf{\method (ours)} & \bestpm{74.1}{0.7} & \bestpm{76.1}{0.7} & \bestpm{79.5}{0.3} & \bestpm{86.5}{0.7} & 89.1\sd{0.5} \\
\midrule
\multicolumn{6}{@{}l}{\textit{FLUX.2-klein}} \\
\rowcolor{basebg}
Base model & 76.4\sdz{0.0} & 79.1\sdz{0.0} & 81.1\sdz{0.0} & 87.6\sdz{0.0} & 90.1\sdz{0.0} \\
Positive teacher & 72.7\sd{1.6} & 77.5\sd{0.7} & 80.7\sd{0.1} & 86.3\sd{0.4} & 89.8\sd{0.3} \\
Cross-author pair & \bestpm{79.6}{1.3} & 79.9\sd{1.1} & 81.6\sd{0.2} & 88.0\sd{0.6} & \bestpm{90.6}{0.2} \\
\rowcolor{oursbg}
\textbf{\method (ours)} & 77.5\sd{1.7} & \bestpm{80.0}{0.5} & \bestpm{82.0}{0.4} & \bestpm{88.7}{0.7} & \bestpm{90.6}{0.4} \\
\bottomrule
\end{fittab}
\end{table*}

\begin{table*}[t]
\centering
\footnotesize
\renewcommand{\arraystretch}{1.04}
\caption{\textbf{Image-quality metrics on held-out GenEval2 prompts.} On the test images from \cref{tab:heldout}, we report mean $\pm$ std across seeds for ImageReward (IR), LAION aesthetics (Aes.), and HPSv2.1 ($\times100$, HPS); single-run entries omit $\pm$. \textbf{Bold} marks the best trained mean per column.}
\label{tab:quality}
\begin{fittab}{\textwidth}{@{}F{l} ccc ccc ccE{c}@{}}
\toprule
& \multicolumn{3}{c}{Qwen-Image-2512} & \multicolumn{3}{c}{SD3.5-M} & \multicolumn{3}{c@{}}{FLUX.2-klein} \\
\cmidrule(lr){2-4}\cmidrule(lr){5-7}\cmidrule(l){8-10}
Method & IR $\uparrow$ & Aes. $\uparrow$ & HPS $\uparrow$ & IR $\uparrow$ & Aes. $\uparrow$ & HPS $\uparrow$ & IR $\uparrow$ & Aes. $\uparrow$ & HPS $\uparrow$ \\
\midrule
\rowcolor{basebg}
Base model & 1.63\sdz{0.00} & 5.95\sdz{0.00} & 34.3\sdz{0.0} & 1.26\sdz{0.00} & 5.58\sdz{0.00} & 30.0\sdz{0.0} & \phantom{$-$}1.20\sdz{0.00} & 5.47\sdz{0.00} & 29.6\sdz{0.0} \\
Positive teacher & 1.61\sd{0.01} & 5.92\sd{0.02} & 33.9\sd{0.1} & 1.34\sd{0.06} & 5.65\sd{0.01} & 30.3\sd{0.3} & \phantom{$-$}1.15\sd{0.04} & \bestpm{5.55}{0.01} & 29.9\sd{0.1} \\
SFT on revised prompt & 1.58\sdz{0.00} & 5.88\sdz{0.00} & 32.7\sdz{0.0} & 1.36\sdz{0.00} & 5.59\sdz{0.00} & \best{31.1}\sdz{0.0} & $-1.84$\sdz{0.00} & 4.05\sdz{0.00} & 17.3\sdz{0.0} \\
DiffusionNFT & 1.63\sd{0.01} & \bestpm{5.93}{0.01} & 34.3\sd{0.2} & 1.27\sd{0.06} & 5.56\sd{0.04} & 29.8\sd{0.7} & \phantom{$-$}1.20\sd{0.02} & 5.49\sd{0.02} & 29.9\sd{0.3} \\
Cross-author pair & 1.64\sd{0.03} & 5.81\sd{0.02} & 32.9\sd{0.2} & \bestpm{1.44}{0.05} & \bestpm{5.67}{0.05} & 30.6\sd{0.3} & \phantom{$-$}1.10\sd{0.26} & 5.54\sd{0.05} & 29.2\sd{1.3} \\
User hindsight & 1.75\sd{0.01} & \bestpm{5.93}{0.04} & \bestpm{35.2}{0.0} & 0.99\sd{0.22} & 5.25\sd{0.02} & 27.8\sd{0.9} & --- & --- & --- \\
\rowcolor{oursbg}
\textbf{\method (ours)} & \bestpm{1.76}{0.00} & \bestpm{5.93}{0.03} & 34.7\sd{0.1} & 1.43\sd{0.10} & 5.45\sd{0.07} & 30.2\sd{0.4} & \phantom{$-$}\bestpm{1.39}{0.03} & 5.50\sd{0.03} & \bestpm{30.2}{0.3} \\
\bottomrule
\end{fittab}
\end{table*}

\begin{table*}[t]
\centering
\footnotesize
\renewcommand{\arraystretch}{1.02}
\caption{\textbf{Zero-shot T2I-CompBench++ evaluation.} Using the official evaluators, we score the seed-1 checkpoints from \cref{tab:heldout} ($\times100$) across all eight validation splits (2,394 images per model). \textbf{Bold} marks the best trained model per backbone.}
\label{tab:t2i}
\begin{fittab}{\textwidth}{@{}F{l} ccc ccc cE{c}@{}}
\toprule
& \multicolumn{3}{c}{Attribute binding} & \multicolumn{3}{c}{Spatial \& numeracy} & \multicolumn{2}{c@{}}{Holistic} \\
\cmidrule(lr){2-4}\cmidrule(lr){5-7}\cmidrule(l){8-9}
Method & Color $\uparrow$ & Shape $\uparrow$ & Texture $\uparrow$ & Spatial $\uparrow$ & 3D-spatial $\uparrow$ & Numeracy $\uparrow$ & Non-spatial $\uparrow$ & Complex $\uparrow$ \\
\midrule
\multicolumn{9}{@{}l}{\textit{Qwen-Image-2512}} \\
\rowcolor{basebg}
Base model & 80.4 & 61.5 & 74.7 & 28.7 & 40.7 & 69.6 & 30.9 & 39.4 \\
Cross-author pair & 82.3 & 62.2 & 74.9 & \best{31.8} & \best{42.4} & 69.7 & \best{31.2} & 39.9 \\
\rowcolor{oursbg}
\textbf{\method (ours)} & \best{84.0} & \best{62.4} & \best{77.0} & 30.4 & 42.1 & \best{71.2} & 31.1 & \best{40.7} \\
\midrule
\multicolumn{9}{@{}l}{\textit{SD3.5-M}} \\
\rowcolor{basebg}
Base model & 79.8 & 55.5 & 72.8 & 30.4 & 37.1 & 60.1 & 31.6 & 37.3 \\
Cross-author pair & 83.5 & 61.1 & 76.3 & \best{34.9} & 39.9 & 63.4 & \best{31.8} & 38.6 \\
\rowcolor{oursbg}
\textbf{\method (ours)} & \best{84.5} & \best{63.3} & \best{77.3} & 32.2 & \best{40.2} & \best{64.5} & 31.6 & \best{38.8} \\
\midrule
\multicolumn{9}{@{}l}{\textit{FLUX.2-klein}} \\
\rowcolor{basebg}
Base model & 79.7 & 55.4 & 72.6 & 35.0 & 38.2 & 57.5 & 31.0 & 38.5 \\
Cross-author pair & 82.4 & 56.1 & \best{74.5} & \best{39.1} & 37.7 & \best{63.1} & \best{31.4} & 38.4 \\
\rowcolor{oursbg}
\textbf{\method (ours)} & \best{83.4} & \best{58.3} & 73.7 & 38.8 & \best{37.9} & 61.0 & 30.9 & \best{38.7} \\
\bottomrule
\end{fittab}
\end{table*}

\begin{table*}[t]
\centering
\footnotesize
\renewcommand{\arraystretch}{1.02}
\caption{\textbf{OneIG-Bench under the official Qwen2.5-VL-7B scorer.} On the test images from \cref{tab:ood}, we report mean $\pm$ std across seeds ($\times100$) for OneIG-Bench's official General-Object (Obj.) and Portrait (Por.) scores. \textbf{Bold} marks the best mean per column.}
\label{tab:oneig_official}
\begin{fittab}{\textwidth}{@{}F{l} cc cc cE{c}@{}}
\toprule
& \multicolumn{2}{c}{Qwen-Image-2512} & \multicolumn{2}{c}{SD3.5-M} & \multicolumn{2}{c@{}}{FLUX.2-klein} \\
\cmidrule(lr){2-3}\cmidrule(lr){4-5}\cmidrule(l){6-7}
Method & Obj. $\uparrow$ & Por. $\uparrow$ & Obj. $\uparrow$ & Por. $\uparrow$ & Obj. $\uparrow$ & Por. $\uparrow$ \\
\midrule
\rowcolor{basebg}
Base model & 83.5\sdz{0.0} & 88.2\sdz{0.0} & 72.1\sdz{0.0} & 79.0\sdz{0.0} & 77.4\sdz{0.0} & 83.3\sdz{0.0} \\
Positive teacher & 83.1\sd{0.1} & 88.4\sd{0.1} & 72.9\sd{0.8} & 80.2\sd{0.3} & 75.8\sd{0.5} & 82.7\sd{0.3} \\
Cross-author pair & 83.7\sd{0.2} & \bestpm{88.6}{0.3} & 74.7\sd{0.2} & \bestpm{80.8}{0.2} & 78.1\sd{0.9} & 83.9\sd{0.4} \\
User hindsight & 83.1\sd{0.0} & 87.7\sd{0.3} & 70.9\sd{2.2} & 78.1\sd{1.5} & --- & --- \\
\rowcolor{oursbg}
\textbf{\method (ours)} & \bestpm{84.0}{0.0} & 88.4\sd{0.2} & \bestpm{75.0}{1.1} & 80.4\sd{0.7} & \bestpm{79.0}{0.5} & \bestpm{84.0}{0.3} \\
\bottomrule
\end{fittab}
\end{table*}

\begin{table*}[t]
\centering
\footnotesize
\renewcommand{\arraystretch}{1.04}
\caption{\textbf{Direct training on GenEval with Qwen-Image-2512.} Models train for four rounds on the 553 GenEval prompts at four rollouts per prompt (current weights). We report official Mask2Former overall ($\times100$, All) and per-skill accuracies (\%, with single-object accuracy above 99\% for all methods). \textbf{Bold} marks the best trained model per column.}
\label{tab:geneval}
\begin{fittab}{\textwidth}{@{}F{l} cccccE{c}@{}}
\toprule
Method & All $\uparrow$ & Two-object $\uparrow$ & Counting $\uparrow$ & Color $\uparrow$ & Position $\uparrow$ & Attribute binding $\uparrow$ \\
\midrule
\rowcolor{basebg}
Base model & 73.9 & 92.9 & 60.3 & 86.4 & 46.0 & 57.8 \\
Positive teacher & 74.8 & \best{93.4} & 60.0 & 87.0 & 44.8 & 63.8 \\
Cross-author pair & 74.6 & 92.2 & 60.6 & 87.8 & 45.0 & 62.8 \\
\rowcolor{oursbg}
\textbf{\method (ours)} & \best{75.8} & 92.2 & \best{61.9} & \best{89.9} & \best{45.8} & \best{65.5} \\
\bottomrule
\end{fittab}
\end{table*}

\paragraph{OneIG-Bench, T2I-CompBench++, and direct GenEval.} \cref{tab:ood_full} adds seed standard deviations to \cref{tab:ood} and reports OneIG-Bench's General-Object and Portrait subsets separately under Qwen3-VL-8B and InternVL3.5-38B (J2). \method shows low seed variance (at most 1.7 points on GenEval and 0.7 on OneIG-Bench), beats the cross-author pair on General Object across all three backbones under both judges, and stays within 0.6 points of the base model on Portrait. Scoring the same images with OneIG-Bench's official Qwen2.5-VL-7B judge (\cref{tab:oneig_official}) confirms the same ranking on General Object ($84.0$\sd{0.0}, $75.0$\sd{1.1}, and $79.0$\sd{0.5} for \method vs.\ 83.5, 72.1, and 77.4 base). On zero-shot T2I-CompBench++ (\cref{tab:t2i}), \method improves over every base model on Color, Shape, Texture, 2D Spatial, Numeracy, and Complex; compared with the cross-author pair, it leads on Color, Shape, and Complex across all three backbones (winning eight of the nine attribute-binding scores), stays within 0.5 points on 3D-spatial and Non-spatial, and trails slightly only on 2D Spatial. In \cref{tab:geneval}, we also train directly on GenEval's 553 prompts with Qwen-Image-2512; because single-object accuracy already exceeds 99\% and one rollout per prompt yields few failures, all methods sample four rollouts per prompt. \method raises overall accuracy from 73.9 to 75.8, ahead of the positive self-teacher (74.8) and cross-author pair (74.6).

\paragraph{Image-quality metrics.} \cref{tab:quality} scores the held-out test images from \cref{tab:heldout} with ImageReward~\cite{xu2023imagereward}, LAION aesthetics, and HPSv2.1~\cite{wu2023hpsv2}. Canceling the critic's unrequested scene additions does not hurt visual quality: \method improves both ImageReward (1.63 to 1.76 on Qwen-Image-2512, 1.26 to 1.43 on SD3.5-M, and 1.20 to 1.39 on FLUX.2-klein) and HPSv2.1 over all three base models. It also preserves or slightly improves PickScore (23.86 to 23.99 on Qwen-Image-2512 in \cref{tab:qwen}; 22.88 to 22.98 across seeds vs.\ 22.86 base on SD3.5-M) and keeps LAION aesthetics within 0.13 points of the base models. By contrast, the cross-author pair lowers HPSv2.1 below the base model on Qwen-Image-2512 (32.9 vs.\ 34.3) and FLUX.2-klein (29.2 vs.\ 29.6), and user hindsight degrades all three metrics on SD3.5-M.

\subsection{Additional Ablations and Qualitative Results}

\begin{table*}[t]
\centering
\footnotesize
\renewcommand{\arraystretch}{1.04}
\caption{\textbf{Token-state alignment variants on FLUX.2-klein.} Expanding Rows 9 and 10 of \cref{tab:ablation} under a frozen teacher ($\lambda=2$, six rounds, current weights), we report two-seed mean $\pm$ std under the default critic and single-seed scores under the sunset critic. \textbf{Bold} marks the best GM and Sun.\ rate.\looseness=-1}
\label{tab:embedding}
\begin{fittab}{\textwidth}{@{}F{l} cc cc E{c}@{}}
\toprule
& \multicolumn{2}{c}{Default critic} & \multicolumn{3}{c@{}}{Sunset critic} \\
\cmidrule(lr){2-3}\cmidrule(l){4-6}
Twin encoding & GM $\uparrow$ & Plain (\%) & GM $\uparrow$ & Sun. (\%) $\downarrow$ & Plain (\%) \\
\midrule
\rowcolor{basebg}
Base model & 23.5\sdz{0.0} & 51.4\sdz{0.0} & 23.5 & 2.7 & 51.4 \\
Independent encoding ($r$ and $\rt$) & \bestpm{24.9}{0.3} & 45.2\sd{1.1} & 24.0 & 4.7 & 14.9 \\
Suffix + padding copied & --- & --- & 21.9 & 2.5 & 52.6 \\
\rowcolor{oursbg}
\textbf{Suffix alignment (\method)} & 24.6\sd{0.2} & 56.3\sd{1.7} & \best{25.1} & \best{2.0} & 62.3 \\
\bottomrule
\end{fittab}
\end{table*}

\def\suppanalysispart{1}\ifdefined\suppanalysispart\else\def\suppanalysispart{0}\fi
\ifnum\suppanalysispart<2
\begin{table*}[t]
\centering
\footnotesize
\renewcommand{\arraystretch}{1.08}
\caption{\textbf{Reintroducing rewrite drift on SD3.5-M.} Extending \cref{tab:dose} (\cref{sec:supp_impl}), we report mean $\pm$ std across seeds for Soft-TIFA GM, J2, and habit-probe rates. $^{\ast}$Standard deviation is omitted at $\lambda_{\rm d}=2.0$ because the two default-critic seeds drift toward different habits (detailed scenery vs.\ sunset skies; see text). \textbf{Bold} marks the best trained model.}
\label{tab:dose_full}
\begin{fittab}{\textwidth}{@{}F{l} cccc cccE{c}@{}}
\toprule
& \multicolumn{4}{c}{Default critic (EMA weights)} & \multicolumn{4}{c@{}}{Sunset critic (current weights)} \\
\cmidrule(lr){2-5}\cmidrule(l){6-9}
Setting & GM $\uparrow$ & J2 $\uparrow$ & Plain (\%) & Sun. (\%) & GM $\uparrow$ & J2 $\uparrow$ & Plain (\%) & Sun. (\%) $\downarrow$ \\
\midrule
\rowcolor{basebg}
Base model & 19.6\sdz{0.0} & 35.4\sdz{0.0} & 62.2\sdz{0.0} & \phantom{0}2.2\sdz{0.0} & 19.6\sdz{0.0} & 35.4\sdz{0.0} & 62.2\sdz{0.0} & \phantom{0}2.2\sdz{0.0} \\
\rowcolor{oursbg}
\textbf{$\lambda_{\rm d}=0$ (\method)} & \bestpm{30.7}{1.1} & \bestpm{47.9}{1.2} & 76.2\sd{3.6} & \phantom{0}1.5\sd{0.2} & \bestpm{30.2}{0.5} & \bestpm{47.2}{0.7} & 71.9\sd{7.3} & \phantom{0}\bestpm{2.4}{0.2} \\
$\lambda_{\rm d}=0.5$ & 30.1\sd{0.9} & 47.4\sd{1.5} & 63.6\sd{4.7} & \phantom{0}1.4\sd{0.4} & 28.2\sd{2.3} & 46.4\sd{2.4} & 17.4\sd{0.8} & 95.4\sd{0.0} \\
$\lambda_{\rm d}=1.0$ & 28.9\sd{0.7} & 45.9\sd{0.3} & 46.8\sd{1.7} & \phantom{0}1.8\sd{0.3} & 23.6\sd{0.7} & 42.0\sd{0.6} & 13.1\sd{4.3} & 99.5\sd{0.0} \\
$\lambda_{\rm d}=2.0$ (noise-weighted pair) & 26.5\sd{2.6} & 44.8\sd{3.5} & 53.1\rlap{$^{\ast}$}\sdz{0.0} & 14.2\rlap{$^{\ast}$}\sdz{0.0} & 16.0\sd{2.1} & 33.5\sd{4.3} & \phantom{0}5.4\sd{1.3} & 99.6\sd{0.1} \\
\bottomrule
\end{fittab}
\end{table*}
\fi

\ifnum\suppanalysispart=1\else
\begin{table*}[t]
\centering
\footnotesize
\renewcommand{\arraystretch}{1.08}
\caption{\textbf{Sensitivity to simulated critic errors on SD3.5-M.} Extending \cref{subsec:noise}, each round randomly replaces 0\%, 25\%, or 50\% of valid critiques with critiques from other failed prompts, and we report two-seed mean $\pm$ std under EMA weights after six rounds. \textbf{Bold} marks the higher alignment mean per row.}
\label{tab:noise_full}
\newcolumntype{O}{>{\columncolor{oursbg}[\tabcolsep][\dimexpr\tabcolsep+0.5pt\relax]}c}
\newcolumntype{Q}{>{\columncolor{oursbg}[\dimexpr\tabcolsep+0.5pt\relax][\dimexpr\tabcolsep+0.5pt\relax]}c}
\newcolumntype{P}{>{\columncolor{oursbg}[\dimexpr\tabcolsep+0.5pt\relax][0pt]}c}
\begin{fittab}{\textwidth}{@{}F{l} cccc OQQP@{}}
\toprule
& \multicolumn{4}{c}{Cross-author pair} & \multicolumn{4}{c@{}}{\textbf{\method (ours)}} \\
\cmidrule(lr){2-5}\cmidrule(l){6-9}
Corrupted critiques & GM $\uparrow$ & J2 $\uparrow$ & Plain (\%) & Sun. (\%) & \multicolumn{1}{c}{GM $\uparrow$} & \multicolumn{1}{c}{J2 $\uparrow$} & \multicolumn{1}{c}{Plain (\%)} & \multicolumn{1}{E{c}@{}}{Sun. (\%)} \\
\midrule
\phantom{0}0\% (clean critiques) & 30.2\sd{0.8} & 47.5\sd{0.6} & 54.3\sd{4.8} & 3.3\sd{0.1} & \bestpm{30.9}{0.5} & \bestpm{48.7}{0.7} & 73.8\sd{0.7} & 1.7\sd{0.2} \\
25\% replaced & 29.7\sd{0.5} & 46.6\sd{0.1} & 52.0\sd{10.1} & 3.1\sd{1.2} & \bestpm{30.4}{1.8} & \bestpm{48.0}{1.2} & 72.0\sd{7.0} & 1.7\sd{0.3} \\
50\% replaced & 27.8\sd{1.9} & \bestpm{45.5}{1.8} & 52.5\sd{5.7} & 4.3\sd{1.1} & \bestpm{28.4}{0.8} & 45.4\sd{1.1} & 74.1\sd{2.7} & 1.4\sd{0.0} \\
\bottomrule
\end{fittab}
\end{table*}
\fi

\paragraph{Token-state alignment and per-skill gains.} Expanding Rows 9 and 10 of \cref{tab:ablation}, \cref{tab:embedding} compares three ways to encode $(r,\rt)$ on FLUX.2-klein under a frozen teacher. Encoding $r$ and $\rt$ independently lets the slot edit alter surrounding tokens through self-attention (\cref{subsec:local}), which lowers plain-background rates from 51.4\% (base) to 45.2\% under the default critic and to 14.9\% (with 4.7\% sunset) under the sunset critic. Copying only the shared content tokens after the slot (\emph{Suffix alignment}) blocks this leakage, keeping backgrounds clean (62.3\% plain, 2.0\% sunset) while raising sunset-critic GM from 24.0 to 25.1. If trailing padding states are copied as well (\emph{Suffix + padding copied}), unequal slot lengths shift where padding begins in $r$ versus $\rt$; although sunset leakage stays low (2.5\%), prompt following drops to 21.9 GM. Removing suffix alignment on Qwen-Image-2512 under the sunset critic similarly lowers GM from $49.3$\sd{1.6} to 47.3 and plain backgrounds from $27.1$\sd{3.9}\% to 21.3\%. Separately, \cref{tab:skills_full} breaks down GenEval2 accuracy by skill on Qwen-Image-2512: \method improves attribute, count, and verb accuracy over the base model under both judges, surpassing the cross-author pair on attributes ($+8.7$ points under Soft-TIFA, $+7.4$ under J2) and verbs ($+11.4$ and $+8.0$).\looseness=-1

\paragraph{Frozen-teacher sunset runs and baseline schedules.} In Rows 1 to 3 of \cref{tab:ablation}, the GM column reports default-critic scores while the Sun.\ column comes from separate frozen-teacher runs on SD3.5-M under the sunset critic. Across three seeds on those sunset-critic runs, hindsight twins reach $23.6$\sd{0.9} GM with $2.2$\sd{0.3}\% sunset images, compared with $19.4$\sd{0.7} GM ($82.1$\sd{2.6}\% sunset) for the positive self-teacher and $23.2$\sd{0.1} GM ($81.9$\sd{6.6}\% sunset) for the cross-author pair. Giving the baselines \method's evolving-teacher schedule under the default critic does not close the gap either: refreshing the positive self-teacher each round reaches only $22.6$\sd{0.9} GM across three seeds, and scaling $\Delta_{\rm pos}$ by $\lambda w(\sigma)$ as in \cref{eq:target} collapses GM to 11.4 because teacher updates compound rewrite drift. For the cross-author pair, refreshing the teacher at the start of each round (Row 7 of \cref{tab:ablation}, 29.2 EMA GM and 46.1 J2) performs nearly the same as the step-wise EMA teacher in \cref{tab:heldout} ($29.1$\sd{1.7} EMA GM and $45.8$\sd{2.4} J2).\looseness=-1

\paragraph{Variants of \method's training schedule.} On SD3.5-M, \method's default schedule reaches $30.7$\sd{1.1} EMA GM at round six and $31.4$\sd{1.0} at round nine. Anchoring accepted rollouts to the frozen base model $\theta_0$ rather than the round-start teacher $\bar\theta$ slightly lowers six-round GM to 30.2, as accepted samples pull the student back toward $\theta_0$. At nine rounds, replacing the smooth noise weight $w(\sigma)=2\sigma$ with a hard step $\mathbf{1}[\sigma\ge0.5]$ drops GM to 24.9, and adding the uncontextualized contrast $\Delta_{\rm hind}$ to the target drops it to 24.1. By contrast, decaying the learning rate by 0.85 per round (31.3 GM) or adding 394 training prompts from T2I-CompBench and Pick-a-Pic (31.7 GM) leaves nine-round performance virtually unchanged, showing that the fixed-rate schedule in \cref{alg:hint} is already stable.

\begin{figure*}[t]
  \centering
  \includegraphics[width=\linewidth]{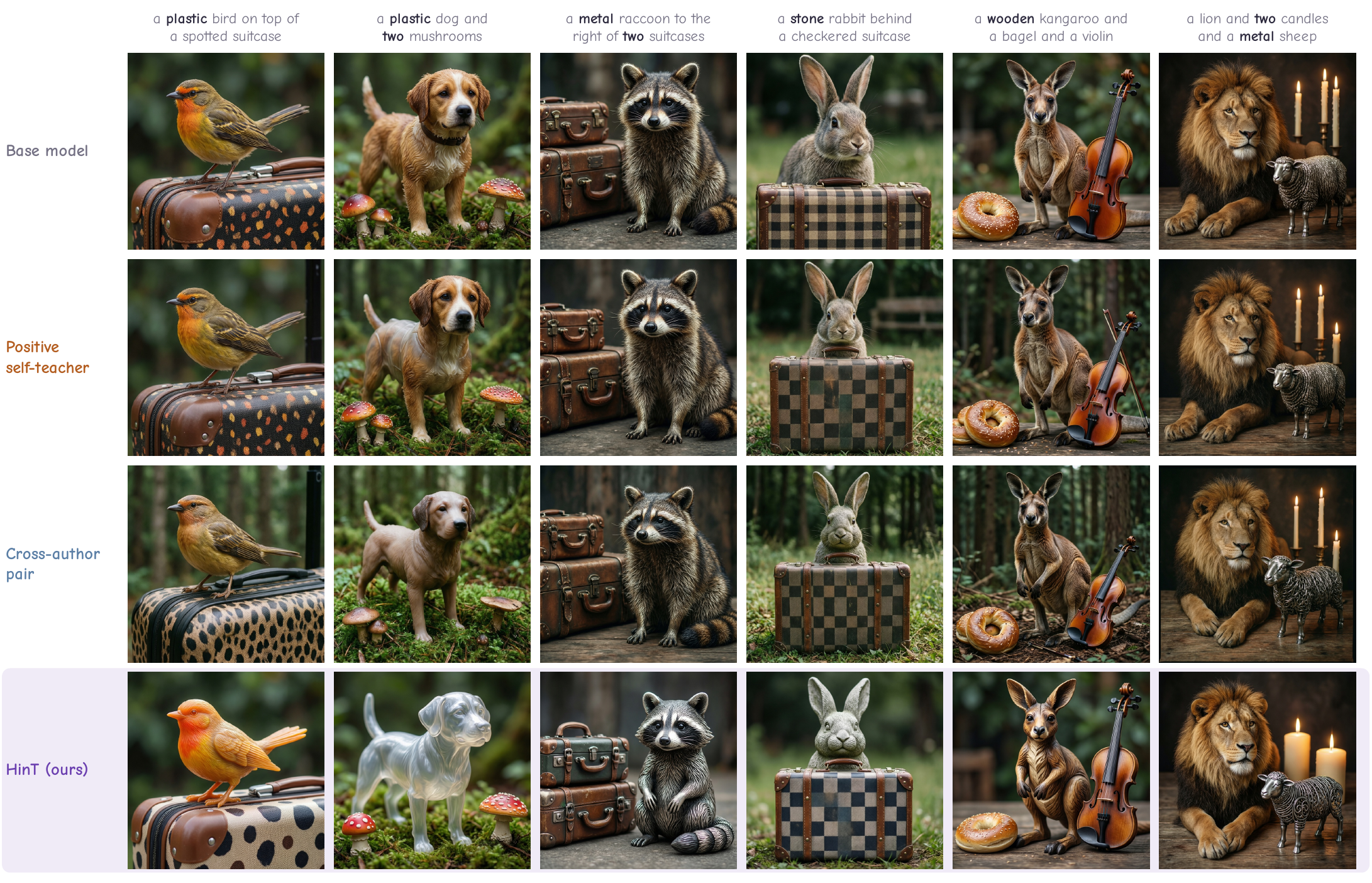}
  \caption{\textbf{Qualitative comparison on Qwen-Image-2512 under the default critic.} Each column shares a prompt and seed: a plastic bird on a spotted suitcase, a plastic dog with two mushrooms, a metal raccoon to the right of two suitcases, a stone rabbit behind a checkered suitcase, a wooden kangaroo with a bagel and a violin, and a lion with two candles and a metal sheep. \method satisfies every GenEval2 question on all six examples under both judges.}
  \label{fig:qualitative}
\end{figure*}

\begin{figure*}[t]
  \centering
  \includegraphics[width=\linewidth]{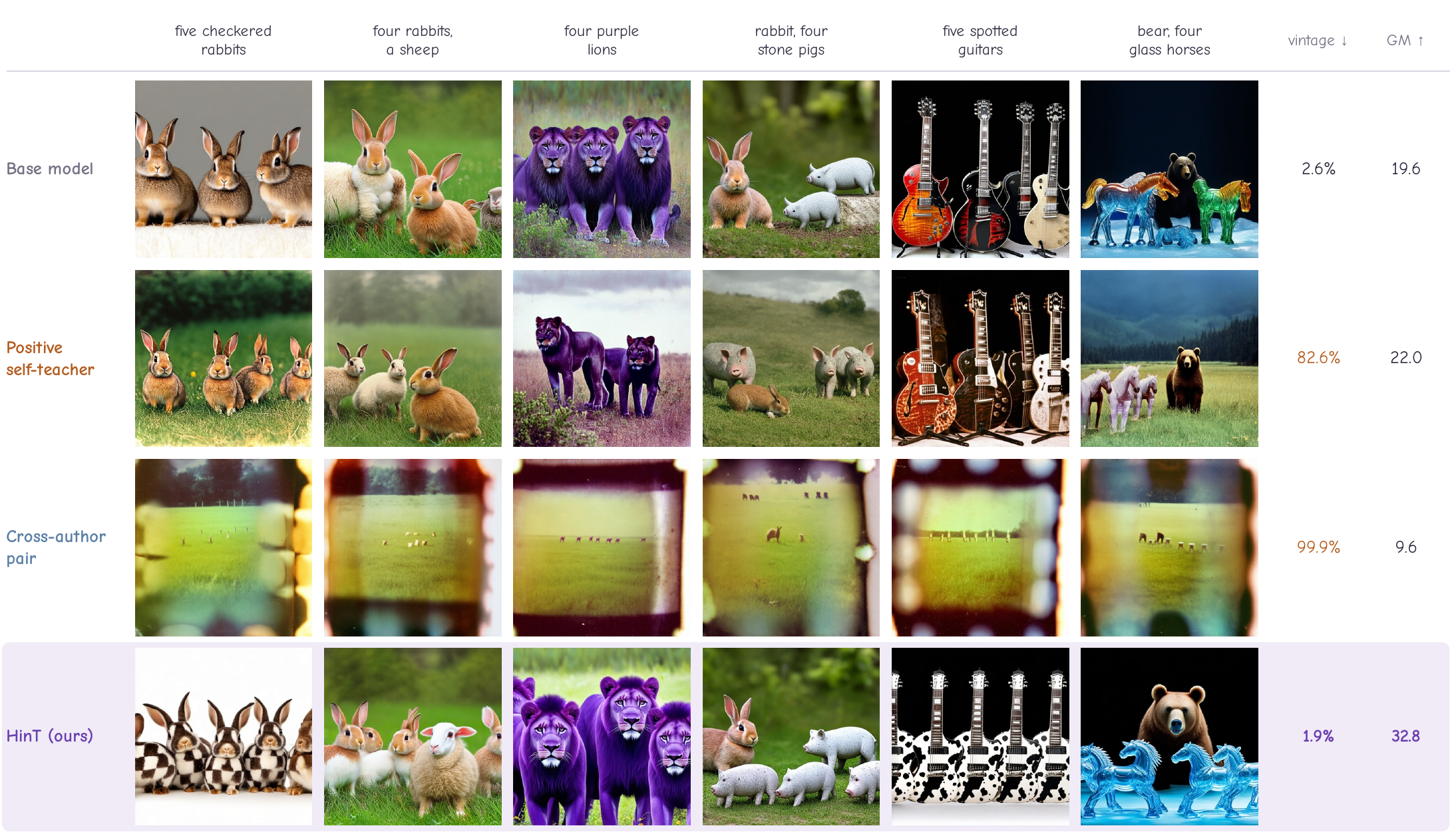}
  \caption{\textbf{Qualitative comparison on SD3.5-M under the vintage-film critic.} Columns share a held-out prompt and seed after six rounds under current weights (rightmost column: seed-1 scores; three-seed means in \cref{tab:habit}). The positive self-teacher fades the color palette, and the cross-author pair turns scenes into distant, grain-bordered meadows where requested objects shrink or vanish (9.6 to 13.1 GM across seeds vs.\ 19.6 base). \method keeps clean rendering (1.9\% to 4.0\% vintage rate) and satisfies the object prompts (28.5 to 32.8 GM).\looseness=-1}
  \label{fig:vintage}
\end{figure*}

\paragraph{Reintroducing drift, critic errors, and qualitative results.} \cref{tab:dose_full,tab:noise_full} report the full seed means, J2 scores, and habit-probe rates for the analyses in \cref{sec:analysis}. In \cref{tab:dose_full}, reintroducing even half-strength rewrite drift ($\lambda_{\rm d}=0.5$) drives sunset prevalence to 95.4\% under the sunset critic; at full drift ($\lambda_{\rm d}=2.0$) under the default critic, the two seeds diverge toward different habits in the critic's rewrites, with one filling in scenic backgrounds (27.7\% plain, 2.5\% sunset) and the other latching onto warm-light descriptions to produce sunset skies on 25.9\% of images (78.5\% plain). In \cref{tab:noise_full}, by contrast, replacing up to 50\% of valid critiques with critiques from other failed prompts lowers alignment only modestly (from 30.9 to 28.4 GM) without inducing style drift (74.1\% plain, 1.4\% sunset), since uncorrelated errors cancel out across prompts whereas systematic rewrite habits accumulate. Finally, \cref{fig:qualitative,fig:vintage} show qualitative comparisons on Qwen-Image-2512 under the default critic and on SD3.5-M under the vintage-film critic.

\newpage
\section{Limitations and Future Work}
\label{sec:supp_lim}

\paragraph{Span-level versus global spatial errors.} \method localizes each correction to a failed phrase inside a shared scene template, which is naturally suited to object attributes, materials, counts, and actions in \cref{tab:skills_full,tab:t2i}. When a failure instead stems from global multi-object layout across the canvas, swapping a relational phrase in text provides weaker geometric guidance than explicit spatial conditioning, explaining why 2D spatial gains in \cref{tab:t2i}, while consistently positive over every base model, are smaller than attribute gains. Combining slotted twin contrasts with lightweight spatial cues such as critic-predicted bounding boxes is a natural next step.\looseness=-1

\paragraph{Training-time critique and extension beyond static images.} While \method adds zero latency or VLM calls at inference time and finishes post-training in under two hours across backbones, each training rollout still queries a 30B-scale open-source VLM once to produce structured diagnoses and slotted templates; distilling slotted template generation into a smaller verifier could further accelerate larger-scale runs. More broadly, VLM critics in instruction-guided image editing and text-to-video generation similarly introduce unrequested camera motion, pacing, or color grading when expanding prompts, making temporal and editing extensions of \method an exciting direction for future work~\cite{liu2025revise,team2026behind}.\looseness=-1

\clearpage
\def\suppanalysispart{2}

\end{document}